\PassOptionsToPackage{dvipsnames, table, xcdraw}{xcolor}
\documentclass{vivo}
\usepackage{booktabs}
\usepackage{amsmath, amssymb, mathtools, bm}
\usepackage{multirow, enumitem, float, wrapfig}
\usepackage{graphicx} % DO NOT CHANGE THIS
\usepackage{newfloat}
\usepackage{listings}
\usepackage{booktabs} % for \toprule, \midrule, \bottomrule
\usepackage{multirow} % for multirow functionality in tables
\usepackage{caption}
\usepackage{subcaption}
\usepackage[numbers]{natbib}
\usepackage{cleveref}
\usepackage{pdfpages}

\PassOptionsToPackage{hyphens}{url} 
\usepackage{url}

\usepackage[utf8]{inputenc}
\usepackage{tabularx}

\newcommand{\nMethod}{SmartPhotoCrafter}

\title{\nMethod{}: Unified Reasoning, Generation and Optimization for Automatic Photographic Image Editing}

\author[]{Ying Zeng$^{\dagger}$} 
\author[]{Miaosen Luo}
\author[]{Guangyuan Li}
\author[]{Yang Yang} 
\author[]{Ruiyang Fan} 
\author[]{Linxiao Shi} 
\author[]{Qirui Yang} 
\author[]{Jian Zhang} 
\author[]{Chengcheng Liu}
\author[]{Siming Zheng} 
\author[]{Jinwei Chen} 
\author[]{Bo Li}
\author[]{Peng-Tao Jiang$^{\dagger}$}

\affiliation[]{vivo BlueImage Lab, vivo Mobile Communication Co., Ltd.} 
\affiliation[]{ $\dagger$: Project lead.}

\abstract{
Traditional photographic image editing typically requires users to possess sufficient aesthetic understanding to provide appropriate instructions for adjusting image quality and camera parameters. 
However, this paradigm relies on explicit human instruction of aesthetic intent, which is often ambiguous, incomplete, or inaccessible to non-expert users. 
Besides, recent editing models mostly rely on user-provided instructions, while lacking the capability to understand aesthetic deficiencies and reason about improvement strategies. 
In this work, we propose \textbf{SmartPhotoCrafter}, an automatic photographic image editing method which formulates image editing as a tightly coupled reasoning-to-generation process. 
The proposed model first performs image quality comprehension and identifies deficiencies by the \textbf{Image Critic} module, and then the \textbf{Photographic Artist} module realizes targeted edits to enhance image appeal, eliminating the need for explicit human instructions. 
A multi-stage training pipeline is adopted: 
(i) Foundation pretraining to establish basic aesthetic understanding and editing capabilities, (ii) Adaptation with reasoning-guided multi-edit supervision to incorporate rich semantic guidance, and (iii) Coordinated reasoning-to generation reinforcement learning to jointly optimize reasoning and generation. During training, SmartPhotoCrafter emphasizes photo-realistic image generation, while supporting both image restoration and retouching tasks with consistent adherence to color- and tone-related semantics.
We also construct a stage-specific dataset, which progressively builds reasoning and controllable generation, effective cross-module collaboration, and ultimately high-quality photographic enhancement.
Experiments demonstrate that SmartPhotoCrafter outperforms existing generative models on the task of automatic photographic enhancement, achieving photo-realistic results while exhibiting higher tonal sensitivity to retouching instructions.
}

\projectpage{https://github.com/vivoCameraResearch/SmartPhotoCrafter}

\begin{document}

\maketitle

\begin{figure}[t]
    \centering
    \includegraphics[page=1, width=\linewidth]{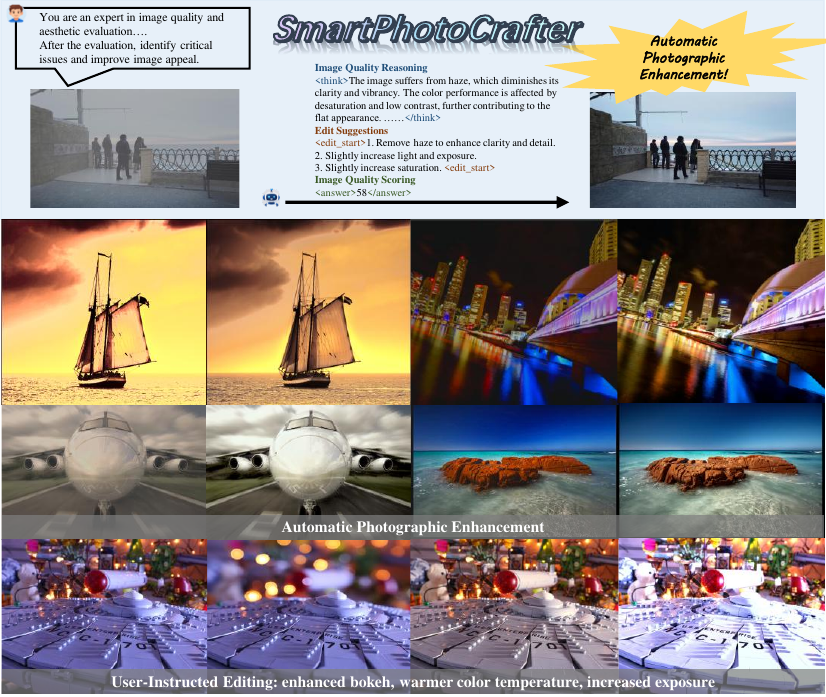}
    \caption{SmartPhotoCrafter achieves automatic photographic image editing within a unified, photographic-aware framework. Beyond user-instructed editing, SmartPhotoCrafter can automatically interpret aesthetic and photometric cues, summarize edit directions, and realize multi-attribute image enhancements.}
\end{figure}

\section{Introduction} \label{sec:intro}

Automatic photographic editing aims to produce high-quality, aesthetically pleasing images while respecting the original semantic content, which is a cognitive process involving aesthetic perception and decision-making to determine appropriate adjustments that enhance visual appeal. Enabling machines to autonomously perform such process remains a fundamental challenge in computational photography and visual intelligence. 
Recent advances in diffusion models and multimodal large language models (MLLMs) have significantly improved the capability of image editing systems. Most existing approaches formulate image editing as an {instruction-conditioned generation task}, where the model modifies an input image according to user-provided textual prompts. 
However, these methods require that the user already knows \textit{what} to edit and \textit{how} to edit, while non-expert users may fail to provide optimal editing instructions, especially for the photographic image enhancement task.
Moreover, instruction-conditioned models lack the ability to perceive image quality and deficiencies, which limits their ability to perform fully automatic image enhancement. 

In this work, we propose SmartPhotoCrafter, a unified reasoning-to-generation framework for automatic photographic editing. SmartPhotoCrafter leverages an MLLM-based Image Critic to condition the Photographic Artist for high-fidelity image editing, achieving representation-level integration between understanding and generation. To enable stable training and semantic-photometric alignment, we design a multi-stage training pipeline: (1) foundation pretraining on image quality assessment (IQA) and image editing datasets to establish reasoning and photographic enhancement skills, (2) reasoning-conditioned adaptation, where the Photographic Artist is conditioned on the Image Critic’s latent reasoning representations to produce reasoning-aware edits, and (3) coordinated reasoning-to-generation reinforcement learning, which jointly optimizes semantic reasoning and photographic editing. In Stage III, the Image Critic and Photographic Artist are trained within a unified reinforcement learning framework: the former refines reasoning and edit suggestions, while the latter generates high-fidelity, semantically consistent edits. The reward is designed for task-oriented photorealistic enhancement, encouraging both modules to explore strategies that improve image appeal. This coordination leads to progressively improved semantic fidelity, edit quality, and photometric precision.

In general, the contributions of this work are as follows:
\begin{itemize}
\item \textbf{Reasoning-to-Generation Unified Framework}: We propose SmartPhotoCrafter, a unified framework that formulates image editing as a tightly coupled reasoning-to-generation process, where the Image Critic conditions the Photographic Artist to produce enhanced image outputs. The representation-level integration bridges semantic understanding and editing execution, enabling outputs that are both interpretable and aesthetically coherent.
\item \textbf{Multi-Stage Training Pipeline}: We design a three-stage training strategy: foundation pretraining for semantic reasoning and editing skills, reasoning-conditioned adaptation for latent-level alignment between the Image Critic and Photographic Artist, and coordinated reinforcement learning for fine-grained, photometrically sensitive optimization.
\item \textbf{Coordinated Reasoning-to-Generation Reinforcement Learning}: We introduce a multi-objective reinforcement learning framework that jointly optimizes the two main modules. {The reward is elaborately designed to enhance attribute-level photometric sensitivity, explicitly improving optimization toward fine-grained tonal adjustments, leading to better alignment with high-quality color distributions.}
\item \textbf{High-Quality Multi-Stage Training Data}:
We construct datasets tailored to different training stages, covering diverse supervision signals for IQA reasoning, edit suggestion, and high-quality photographic targets. This design supports progressive learning of reasoning capability, controllable generation, and cross-module collaboration, ultimately enabling high-quality automatic photographic image editing.

\end{itemize}

\section{Related Works}
Automatic photographic image editing requires models that simultaneously align with high-level semantic editing intentions and preserve low-level photometric fidelity. Existing approaches in this area span automatic image editing, image quality assessment, and image retouching.

\subsection{Automatic Image Editing}
The goal of automatic image editing is to perform image transformations that align with desired semantic changes while minimizing manual intervention. 
The rise of generative models, particularly diffusion-based frameworks~\cite{DDPM,DDIM,LDM,SDXL,flux2024}, has substantially improved the realism and diversity of edits~\cite{camedit,cameramaster,bokehdiffusion}. Large-scale vision-language models have enabled instruction-driven editing towards unified understanding and generation, reducing reliance on dedicated user instructions. 
Several recent works explore this by integrating analysis, reasoning, and editing within {an MLLM-based agent system}~\cite{jarvisevo,jarvisart,agentbanana,photoartagent,photoagent,reasonedit,thinkrl,chang2025pertouch,monetgpt,retouchllm,4kagent}. For example, JarvisEvo~\cite{jarvisevo} extends multimodal large models with agent-style reasoning to perform structured visual tasks. JarvisArt~\cite{jarvisart} introduces a planning-and-execution framework tailored for artistic image manipulation. Agent Banana~\cite{agentbanana} presents a hierarchical agentic planner-executor framework for high-fidelity, object-aware, deliberative editing. However, these methods mainly decompose editing into sequential tool invocations, and their discrete tool-based pipelines leads to higher deployment complexity. In addition, the loosely coupled architecture makes it difficult to achieve unified handling of restoration and enhancement tasks.

Some other works have explored unified image understanding and generation within generative frameworks. For example, Step1X-Edit~\cite{Step1X-Edit} proposes a large instruction-following editing model that unifies multiple editing tasks within a single generative backbone. Similarly, OmniGen2~\cite{omnigen2} adopt multimodal reasoning to analyze user instructions before synthesizing edited outputs. 
Despite their strong semantic reasoning capabilities, these unified generative editors are primarily optimized for high-level semantic transformations, and remain limited in subtle photometric adjustments required for high-quality enhancement. Moreover, existing approaches focus on instruction-driven alignment, lacking a quality-aware reasoning that can diagnose image deficiencies and infer targeted optimization directions, which is essential for automatic photographic enhancement.

\subsection{Image Quality Assessment}
Image quality assessment (IQA) plays a dual role in evaluating image inputs and guiding adjustments for enhancement methods. Before the widespread adoption of multimodal large language models (MLLMs) for IQA, traditional approaches were mainly categorized into full‑reference (FR) and no‑reference (NR) methods. FR methods rely on handcrafted metrics based on prior knowledge of the human visual system~\cite{hore2010image,wang2004image,zhang2011fsim}, while NR methods estimate quality from natural scene statistics~\cite{mittal2012making,mittal2012no}. With the advent of deep learning, CNN‑ and Transformer‑based models~\cite{zhang2018unreasonable,su2020blindly,ke2021musiq} have significantly improved generalization through large‑scale perceptual feature learning.
Recent work has explored leveraging MLLMs for IQA, aiming to move beyond regression‑based perceptual predictors toward reasoning‑aware quality modeling~\cite{q-align,q-insight,VisualQuality,q-instruct,cai2025q,deqa}. MLLMs integrate visual and linguistic understanding, offering a unified framework for perception, generalization, and interpretability in IQA. For example, Q‑Align~\cite{q-align} aligns vision‑language representations with human quality judgments, enabling quality prediction through multimodal semantic alignment. Q‑Insight~\cite{q-insight} introduces instruction‑driven quality analysis, where the model provides structured explanations alongside quality scores. Q‑Ponder~\cite{cai2025q} first distills reasoning paths and then jointly optimizes scoring accuracy and reasoning consistency using group relative policy optimization (GRPO), thereby enhancing interpretability and accuracy. VisualQuality‑R1~\cite{VisualQuality} adopts a GRPO‑based “learning‑to‑rank” approach, which computes relative comparison probabilities between image pairs and uses continuous fidelity as a reward to improve robustness across diverse distortion types.

However, current IQA approaches are often limited in the sensitivity to subtle photometric deviations introduced by complex editing, and are not typically integrated as photographic‑aware training objectives in editing pipelines. This limitation calls for evaluation frameworks that jointly measure semantic consistency and low-level photometric fidelity.

\subsection{Image Retouching}
Image retouching aims to enhance the aesthetic and technical quality of images through photometric adjustments. Traditional learning-based methods~\cite{rule1,rule2,rule3} rely on hand-crafted features to enhance image aesthetic quality, but they are often limited in their ability to generalize across diverse scenes and lighting conditions.
Instead, deep-learning-based methods~\cite{deep1,deep2,deep3} present higher capacity and robustness.
To further address individual preferences, recent frameworks have shifted toward personalized retouching to adapt to specific user styles~\cite{retouching1,retouching2}. These methods predominantly optimize for holistic visual aesthetics rather than instruction compliance, and often lack robust mechanisms to enforce high-level semantic objectives specified by users.
To bridge this gap, Agent-based retouching methods have emerged as a promising frontier~\cite{photoartagent,retouchiq,retouchllm,photoagent}. By leveraging Visual Language Models (VLMs) as the `planner', these agents can decompose complex user instructions into a sequence of interpretable editing steps, allowing for more precise semantic control compared to `black-box' models. For example, PhotoArtAgent~\cite{photoartagent} combines VLMs with advanced natural language reasoning to emulate the creative process. RetouchIQ~\cite{retouchiq} interprets user-specified editing intentions and generates corresponding image adjustments. 
Despite their potential, existing agent frameworks often struggle to simultaneously handle photographic restoration and retouching, failing to achieve a unified synergy between structural recovery and photometric refinement.

To sum up, existing methods typically exhibit a trade-off: they either focus on semantic transformations that lack precise photometric control, or on fine-grained adjustments that struggle to balance global aesthetic enhancement with the restoration requirements.
In contrast, we propose a unified reasoning-to-generation framework that couples MLLM-based aesthetic reasoning with high-fidelity image generation. Leveraging multi-stage training and retouching-aware coordinated reinforcement learning, our approach enables automatic photorealistic edits that are both semantically guided and photometrically sensitive across diverse photographic scenarios.

\section{Methodology}
\subsection{Problem Formalization}
Let $\mathbf{X}$ denote an input image, which may contain various distortions or suboptimal aesthetic attributes, and $\mathbf{X}_{gt}$ denote its corresponding restored or retouched reference. We formulate the task as a reasoning-guided image enhancement problem. The goal of SmartPhotoCrafter is to understand visual quality, perform interpretable aesthetic reasoning, and produce actionable editing instructions, ultimately generating a high-quality output image $\mathbf{X}_e$ that is both visually appealing and consistent with the reference in content and structure.

The learning framework is built upon two complementary but task-aligned modules:
\begin{enumerate}
    \item \textbf{Image Critic} $f_c$, which interprets an input image $\mathbf{X}$ into a structured output comprising three components, including chain-of-thought aesthetic reasoning statement $\mathcal{R}$ that captures semantic and photometric quality cues, edit suggestions $\mathcal{E}$ that specify actionable image transformations, and the predicted score $\mathcal{S}$ reflecting image quality.
    \item \textbf{Photographic Artist} $f_a$, which generates an edited image $\mathbf{X}_e = f_a(\mathbf{X}, \mathbf{H}_c)$ conditioned on the input image $\mathbf{X}$ and the reasoning latent $\mathbf{H}_c$ derived from the Image Critic. This ensures that the generated edits are \emph{semantically consistent} with the reasoning and \emph{photometrically precise} according to the desired retouching objectives.
\end{enumerate}

Overall, SmartPhotoCrafter formulates {automatic photographic enhancement as a unified problem of aesthetic understanding and high-fidelity editing}, where the Image Critic provides interpretable guidance and the Photographic Artist implements fine-grained and controllable editing.

\subsection{Multi-Stage Training Pipeline}
To enable unified aesthetic understanding and faithful photographic image editing, we adopt a multi-stage training pipeline that learns interpretable reasoning, fine-grained photometric sensitivity, and high-quality image enhancement under photographic-aware supervision.

\subsubsection{Stage I: Foundation Pre-training}
In this stage, we establish basic capabilities for each module through supervised fine-tuning (SFT). 
The Image Critic is trained on two types of complementary datasets, including the IQA datasets 
and editing datasets.
The IQA datasets contain images with diverse distortion types and quality levels, along with ground-truth mean opinion scores (MOS), while the editing datasets provide diverse restoration/retouching examples.
For each image, we utilize Qwen2.5-VL-72B~\cite{Qwen2.5-VL} to generate 
a triplet of analysis that contains image quality understanding, edit suggestions, and quality scores, and the 
predicted scores are curated if the ground-truth ones are available.
We perform SFT on Qwen2.5-VL-7B, while masking the score loss for editing samples to avoid interfering with the scoring ability.
The Photographic Artist is trained on large-scale restoration and retouching datasets, including both single-edit (restoration-only and retouching-only) and multi-edit (combined restoration and retouching) tasks. We perform SFT on Qwen-Image-Edit~\cite{qwenimage} with the flow matching objective.

\subsubsection{Stage II: Reasoning-Conditioned Adaptation}
The second stage semantically couples the Image Critic and Photographic Artist, {forming a unified framework through representation-level integration}. During foundation pre-training, the Photographic Artist is trained with simple instruction prompts to acquire basic restoration and retouching skills, {but does not align well with the rich reasoning feedback from the Image Critic to guide automatic image enhancement.}
In this stage, we adapt the Photographic Artist to reasoning-guided editing by conditioning it on the reasoning-based latent states generated by the Image Critic:
\begin{equation}
\mathbf{X}_e =
{f_a}(\mathbf{X}, \mathbf{H}_c), \quad \mathbf{H}_c = \mathrm{Concat}\!\left(
\mathbf{h}_0^{(L)},\mathbf{h}_1^{(L)},\dots, \mathbf{h}_{T-1}^{(L)}
\right),
\end{equation}
where $\mathbf{X}$ denotes the input image, and $\mathbf{H}_c$ denotes the reasoning-based latent produced by $f_c$. 
$T$ is the total number of tokens including both input context tokens and generated reasoning tokens. 
$\mathbf{h}_t^{(L)}$ denotes the hidden representation at step $t$ of the last layer $L$. 
The reasoning-aware latent representation $\mathbf{H}_c$ is obtained by concatenating the context hidden states and reasoning hidden states, and is used as the conditioning signal for generating the enhanced image $\mathbf{X}_e$.

In this stage, we perform SFT on the Photographic Artist.
By conditioning the editing process on the reasoning-aware representations, the model moves beyond simple instruction following and learns to produce semantically grounded edits, which is essential for reasoning-based automatic enhancement. 

\subsubsection{Stage III:  Coordinated Reasoning-to-Generation Reinforcement Learning}
{
While the previous stages establishes basic capabilities, supervised retouching data alone cannot cover the full exploration space of photographic adjustments, and independently training the Critic and Artist lacks closed-loop optimization. To address this, we propose a unified RL framework, where GRPO~\cite{grpo} optimizes the Image Critic (VLM) for discrete reasoning, and DiffusionNFT~\cite{zheng2025diffusionnft} optimizes the Photographic Artist (DiT) for continuous generation.
}

{For the Image Critic, we employ GRPO to refine its reasoning chains, leveraging its critic-free design to reduce computational overhead. Given an input $\mathbf{X}$, a group of $G$ reasoning outputs $\{o_i\}_{i=1}^G$ is sampled from the current policy $\pi_{\theta_{\text{old}}}$. Each output $o_i$ is assigned a reward $r_i$ by a joint reward function. The group-normalized advantage is computed as:}

{
\begin{equation}
A_i = \frac{r_i - \mu_r}{\sigma_r},
\end{equation}
}

{where $\mu_r$ and $\sigma_r$ denote the mean and standard deviation of rewards within the group. The GRPO optimization objective is defined as:}
{
\begin{equation}
\mathcal{L}_{\text{Critic}}(\theta) = \mathbb{E}_{\mathbf{X}, \{o_i\}_{i=1}^G} \left[ \frac{1}{G} \sum_{i=1}^{G} \min\left( \rho_i(\theta) A_i,\, \mathrm{clip}(\rho_i(\theta), 1-\epsilon, 1+\epsilon) A_i \right) - \beta \cdot \mathbb{D}_{\text{KL}}[\pi_\theta \| \pi_{\text{ref}}] \right],
\end{equation}
}

where $\rho_i(\theta)$ is the importance sampling ratio, $\epsilon$ is the clipping threshold, and $\beta$ controls the KL regularization strength.

For the Photographic Artist, we employ DiffusionNFT to align the Artist with human aesthetic preferences. Building upon GRPO, DiffusionNFT extends the group-relative paradigm to the continuous flow domain by reformulating policy alignment as a contrastive interpolation within the velocity field. The DiffusionNFT optimization objective is defined as:

{
\begin{equation}
\mathcal{L}_\text{Artist}(\theta)=\mathbb{E}_{\mathbf{c},\pi^{\text{old}}(\mathbf{x}_0|\mathbf{c}),t}\; p\|\mathbf{v}_{\theta}^{+}(\mathbf{x}_t,\mathbf{c},t)-\mathbf{v}\|_2^2+(1-p)\|\mathbf{v}_{\theta}^{-}(\mathbf{x}_t,\mathbf{c},t)-\mathbf{v}\|_2^2,
\end{equation}
}

where $p := p(o=1 \mid \mathbf{x}_0, \mathbf{c}) \in [0,1]$ is the optimality probability, mapped from a raw reward signal.
$c$ is the conditioning signal that guides the generative process. $\mathbf{v}_{\theta}^{+}(\mathbf{x}_t,\mathbf{c},t)$ and $\mathbf{v}_{\theta}^{-}(\mathbf{x}_t,\mathbf{c},t)$ are the implicit positive and negative velocity fields, defined as:

{
\begin{equation}
\mathbf{v}_{\theta}^{+}(\mathbf{x}_t,\mathbf{c},t) := (1-\beta)\mathbf{v}^{\text{old}}(\mathbf{x}_t,\mathbf{c},t) + \beta\mathbf{v}_{\theta}(\mathbf{x}_t,\mathbf{c},t),\quad
\end{equation}
}

{
\begin{equation}
\mathbf{v}_{\theta}^{-}(\mathbf{x}_t,\mathbf{c},t) := (1+\beta)\mathbf{v}^{\text{old}}(\mathbf{x}_t,\mathbf{c},t) - \beta\mathbf{v}_{\theta}(\mathbf{x}_t,\mathbf{c},t),
\end{equation}
}

where $\beta$ is a guidance hyperparameter. These guided velocity fields shift the policy toward high-reward regions and away from low-reward trajectories, respectively.

\begin{figure}[t]
    \centering
    \includegraphics[page=2, width=\linewidth]{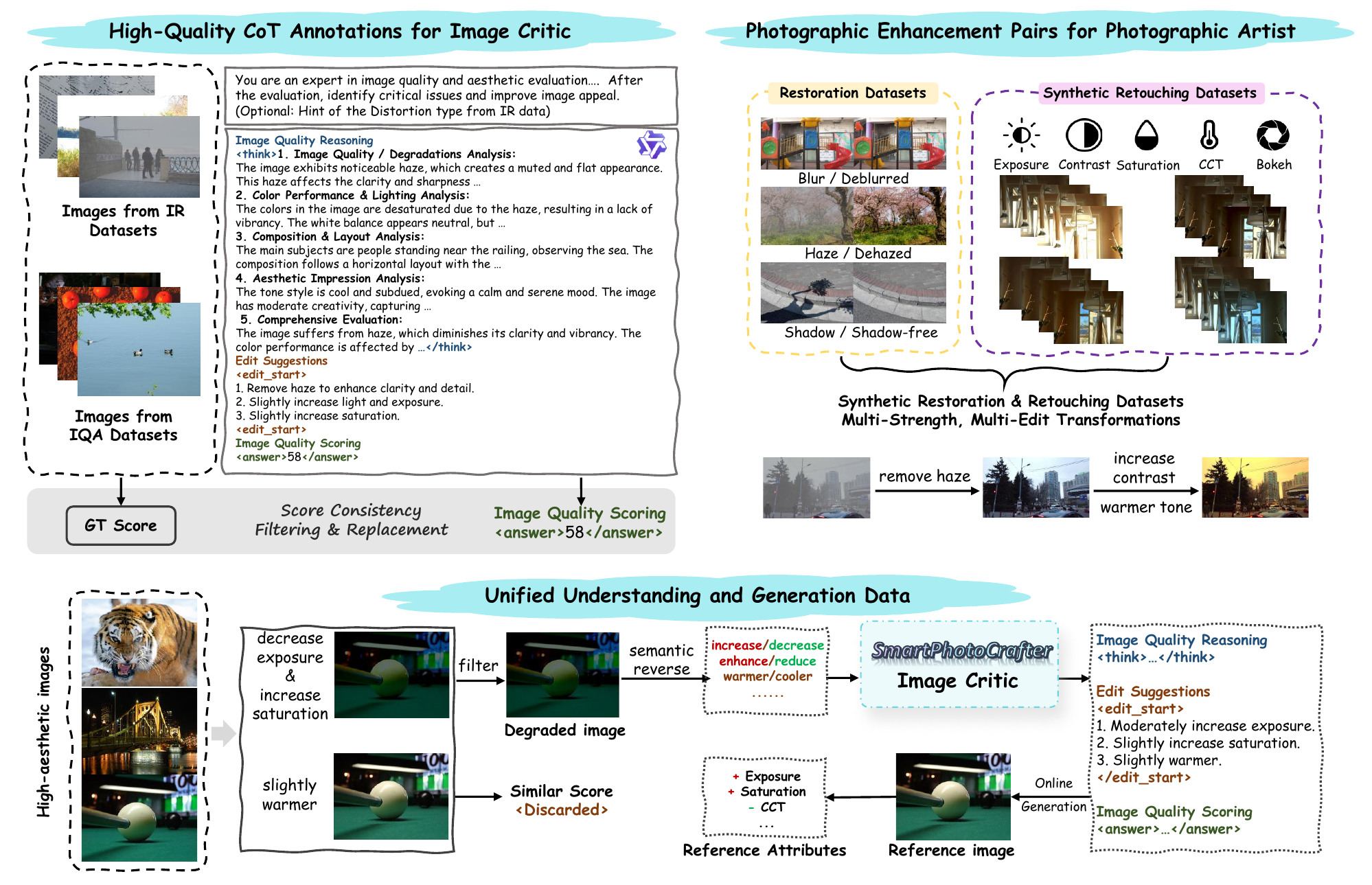}
    \caption{Data generation pipeline of SmartImageCrafter: 1) Annotation of high-quality CoT of image quality reasoning, image scoring and edit suggestions, 2) Generation of photographic enhancement pairs covering image restoration, image retouching and combined scenes, and 3) Creation of unified understanding and generation data for the optimization of both modules.} \label{data_construct}
\end{figure}

\subsection{Data Construction} 
Since each stage optimizes distinct objectives, we design a stage-specific dataset to progressively shape reasoning capability, controllable generation, and cross-module collaboration.
The holistic pipeline of data construction is shown in Figure~\ref{data_construct}.

\subsubsection{Data Construction for Image Critic} 
For Image Critic pretraining, to enable fine-grained reasoning and actionable suggestion generation, we combine IQA datasets (KonIQ~\cite{koniq}, KADID~\cite{kadid} and SPAQ~\cite{spaq}) with multiple editing-related datasets (e.g., GoPro~\cite{gopro}, FoundIR~\cite{foundir}, ISTD~\cite{istd}, etc.), covering various distortion types. 
For each image, we employ Qwen2.5-VL-72B~\cite{Qwen2.5-VL} to generate high-quality chain-of-thought (CoT) reasoning, scalar quality assessment, and structured edit suggestions. 
For IQA datasets, we retain the original MOS scores as ground-truth supervision, and replace model-predicted scores to mitigate regression noise, while the generated reasoning and edit suggestions are preserved. 
For editing-related datasets, distortion categories are directly obtained from dataset annotations. We further exclude samples where the generated reasoning is inconsistent with the annotated distortion labels to ensure supervision reliability.
In addition to restoration datasets, we incorporate bokeh-oriented retouching data, encouraging the model to understand scene-dependent suitability of background blur.
The Image Critic is then trained in an autoregressive SFT manner to generate structured output in the form of [reasoning → suggestion → score], encouraging interpretable image quality understanding and actionable suggestions for further enhancement. 

\subsubsection{Data Construction for Photographic Artist} 
During Stage I, the Photographic Artist is trained to perform distortion removal 
and controllable retouching. 
For restoration tasks, we directly adopt the input–GT pairs from public datasets and construct 
task-specific prompts based on the degradation type (e.g., “remove blur”, “remove haze”, “remove moire”), which enables explicit distortion removal capability.
For image retouching, we construct a synthetic dataset covering exposure, contrast, saturation, correlated color temperature (CCT), and bokeh variations. 
We leverage images from the FilmSet~\cite{filmset} due to their diverse scenes and realistic photographic content, and apply parameterized color and tonal adjustments at multiple intensity levels to generate retouching pairs.
For depth-of-field editing, we leverage multi-level blur pairs from RealBokeh~\cite{realbokeh}, where images with varying blur intensities are directly available. In addition, we select all-in-focus images from BokehDiff~\cite{bokehdiff} and synthesize different blur strengths using a pretrained bokeh model to construct paired data, which provides a multi-level blur space for depth-of-field control.
To enhance compositional editing capability, we further construct multi-edit samples by stacking random retouching adjustments on restoration datasets, forming restoration$+$retouching pairs. This expands the edit space from single-attribute manipulation to structured multi-attribute transformations, enabling the Photographic Artist to model joint editing operators.

\subsubsection{Data Construction for Unified Understanding and Generation}
While the foundation stage trains the two modules independently, the second stage aims at reasoning-conditioned adaptation, which aligns the Image Critic’s latent representations with the Photographic Artist’s conditional generation process.
In this stage, we adopt the editing-related datasets (restoration and retouching) and only finetune the Photographic Artist. 
In addition to the previously adopted data, we incorporate two color grading datasets: the public MIT-Adobe FiveK Dataset~\cite{fivek} and a self-collected subset derived from AVA Dataset~\cite{ava}. 
For the AVA subset, we select high-MOS images as aesthetic references and construct degraded–GT pairs by applying synthetic retouching perturbations, including exposure shifts, contrast reduction, saturation distortion, and color temperature shifts.

{To bridge the gap between understanding and generation, we implement an online generation strategy that dynamically synthesizes reference images. Specifically, the Image Critic analyzes an input image to produce structured editing instructions, which are then applied via the corresponding retouching operations to generate a dynamic target. 
For example, if Image Critic suggests a slight increase in exposure, the reference is constructed by applying the corresponding exposure adjustment to the input image.
The Photographic Artist is subsequently trained to reconstruct this target, conditioned directly on the Image Critic’s latent features. Crucially, this stage enforces alignment at the representation level, where the latent features from the Image Critic serve as semantically grounded editing conditions rather than textual input prompts. Consequently, the training objective shifts from mastering independent capabilities to ensuring cross-module consistency.}

\begin{figure}
    \centering
    \includegraphics[width=\linewidth]{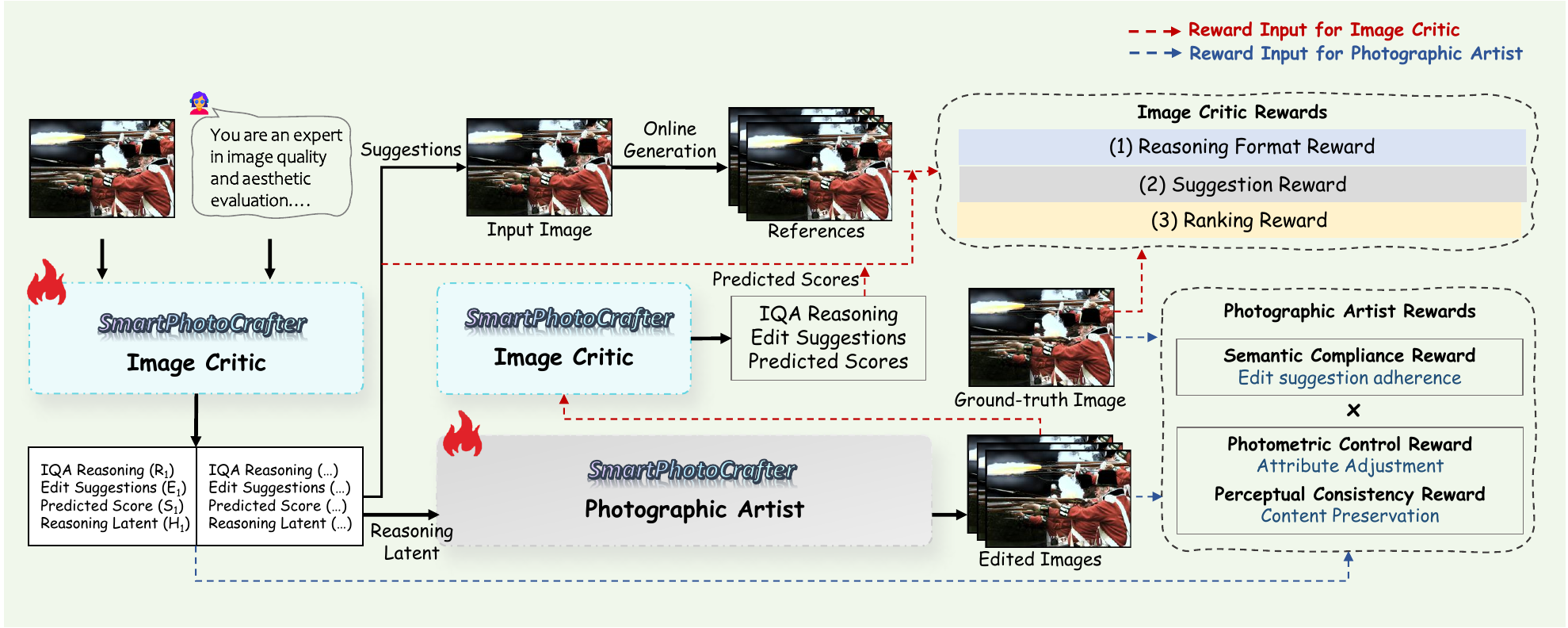}
    \caption{Coordinated reasoning-to-generation reinforcement learning framework of SmartPhotoCrafter. A unified optimization paradigm is employed to jointly enhance the Image Critic and Photographic Artist, enabling photographic-aware reasoning and image enhancement.}
    \label{reward_design}
\end{figure}

\subsection{Reward Design for Photographic-aware Editing}
To optimize the Photographic Artist for high-quality photographic enhancement, as shown in Figure~\ref{reward_design}, we design a photographic-aware multi-level reward mechanism that simultaneously considers semantic-level compliance, photometric-level fidelity, and perceptual-level similarity. 
Existing RL-based methods~\cite{li2025uniworldv2,thinkrl,got-r1,gu2025multireward} often rely on VLM-based perceptual rewards, which provide holistic scores but remain insensitive to subtle photometric variations (e.g., slight exposure or contrast adjustments). To address this, we leverage the structured reasoning outputs from the Image Critic to decompose the supervision signal into distinct dimensions. The overall reward is defined as:
\begin{equation}
    r_{\mathrm{PA}} = r_{\mathrm{comp}}\times (\lambda_1r_{\mathrm{photo}}+\lambda_2r_{\mathrm{perc}}),
\end{equation}
where $r_{\mathrm{comp}}$, $r_{\mathrm{photo}}$, and $r_{\mathrm{perc}}$ denote the Semantic Compliance Reward, Photometric Control Reward, and Perceptual Consistency Reward. $r_{\mathrm{comp}}$ serves as a gating factor that enforces semantic compliance with the editing intentions of the Image Critic. If the prescribed operations are not correctly executed, $r_{\mathrm{comp}}$ is reduced, suppressing the overall reward regardless of the photometric precision. Meanwhile, $r_{\mathrm{photo}}$ provides fine-grained photometric supervision to regulate the magnitude of image adjustments, and $r_{\mathrm{perc}}$ measures LPIPS-based structural and textural fidelity against the ground truth image $\mathbf{X}{gt}$. The $\lambda_1$ and $\lambda_2$ balance the varying optimization difficulties of each component.
We empirically set $\lambda_1$ and $\lambda_2$ to 1.0 and 0.5, respectively.
This coupling ensures that instruction faithfulness is a prerequisite for high rewards, effectively balancing semantic alignment, photometric precision, and perceptual similarity.

\paragraph{Semantic Compliance Reward}
The Semantic Compliance Reward explicitly measures whether the generated image follows the color- and tone-related suggestions summarized by the Image Critic.
Given a set of suggestions $\mathcal{E}=\{e_i\}_{i=1}^{N}$ derived from the Image Critic, we focus our evaluation on the subset of color- and tone-related attributes, and assess whether these photometric adjustments are consistently reflected in the edited image $\mathbf{X}_e$ relative to the input image $\mathbf{X}$. 
By explicitly isolating this subspace, the reward provides a more targeted and reliable supervision signal to improve the compliance with aesthetic retouching intents.

For each suggestion $e_i$, we define a binary compliance indicator:
\begin{equation}
    c_i =
    \begin{cases}
    1, & \text{if the corresponding attribute change is satisfied} \\
    0, & \text{otherwise}
    \end{cases}
\end{equation}

The semantic compliance reward is then computed as the normalized aggregation:
\begin{equation}
    r_{\mathrm{comp}} = \frac{1}{N} \sum_{i=1}^{N} c_i.
\end{equation}
This formulation provides an explicit and disentangled supervision signal, ensuring that each predicted editing operation is verifiably executed. It directly captures controllable attribute-level changes and the model is guided to be more sensitive to subtle retouching behaviors.

By grounding supervision in the reasoning outputs of the Image Critic, this reward enforces consistency between analysis and generation, effectively aligning the Photographic Artist with editing intents.
However, while $r_{\mathrm{comp}}$ ensures correct operation direction, it does not constrain the magnitude or perceptual quality of the enhancement. 
Therefore, we further introduce a photometric control reward and a perceptual consistency reward to supervise fine-grained visual fidelity.

\paragraph{Photometric Control Reward}
The Photometric Control Reward measures the discrepancy in global photometric statistics, rather than enforcing rigid pixel-level reconstruction or global feature matching.
Existing metrics, including feature-space distances or perceptual color differences, collapse all photometric factors into a single scalar signal, leading to ambiguous credit assignment and encouraging shortcut solutions where different attributes may entangle with each other.
To address this limitation, we decompose the photometric transformation into a set of $K$ disentangled attribute functions ${a_k(\cdot)}$, 
where each $a_k$ maps an input image to a scalar that quantifies a specific retouching attribute (e.g., exposure, contrast, saturation or color temperature). These attribute functions are designed to be interpretable and computable in a predefined color space. For example, the exposure attribute can be defined based on the luminance channel in the CIE Lab color space. To characterize attribute-level differences, we measure the deviation of the input image and edited image from the ground truth in each attribute dimension:
\begin{equation}
\Delta a_k^{e} = a_k(\mathbf{X}_e) - a_k(\mathbf{X}), \quad
\Delta a_k^{gt} = a_k(\mathbf{X}_{gt}) - a_k(\mathbf{X}).
\end{equation}
Based on this formulation, supervision is imposed by encouraging the edited image to reduce its attribute-wise deviation from the ground truth. 
In other words, the objective measures whether each attribute in $\mathbf{X}_e$ moves closer to the reference relative to $\mathbf{X}$, which measures relative enhancement rather than enforcing exact matching to the reference or pixel-wise similarity.
This design encourages the model to focus on minimizing residual discrepancies in each attribute dimension, leading to more accurate and interpretable retouching.

However, directly aligning attribute variations may lead to unstable supervision across different editing magnitudes, particularly when the ground-truth adjustment is negligible. To obtain a more stable and interpretable signal, we instead measure the attribute-wise improvement relative to the ground truth. Specifically, for each attribute dimension, the score is computed based on the distance between the edited image and the ground truth in the attribute space, compared against the corresponding distance between the input image and the ground truth. Under this formulation, an optimal edit yields zero distance to the ground truth and thus achieves the highest reward, while suboptimal edits are penalized proportionally to their residual discrepancy. The final reward is obtained by aggregating the attribute-wise scores across all dimensions:
\begin{equation}
r_{\mathrm{attr}}^k =
\begin{cases}
1, & \text{if } |\Delta a_k^{gt}| < \tau_k \ \text{and } |\Delta a_k^{e}| \le \tau_k \\
0, & \text{if } |\Delta a_k^{gt}| < \tau_k \ \text{and } |\Delta a_k^{e}| > \tau_k \\
\dfrac{|\Delta a_k^{e} - \Delta a_k^{gt}|}{|\Delta a_k^{gt}| + \epsilon}, & \text{otherwise}
\end{cases}
\end{equation}
\begin{equation}
r_{\mathrm{photo}} = \frac{1}{K} \sum_{k=1}^K r_{\mathrm{attr}}^k.
\end{equation}
where $\tau_k$ controls the tolerance for negligible ground-truth deviations, $\epsilon$ is a small constant for numerical stability, and $K$ denotes the number of attributes. The final reward is obtained by averaging the attribute-wise scores across all dimensions, reflecting the overall degree to which the edited image reduces its attribute-level discrepancy to the ground truth.

\paragraph{Perceptual Consistency Reward}
The Perceptual Consistency Reward enforces perceptual consistency between the edited image $\mathbf{X}_e$ and the ground-truth image $\mathbf{X}_{gt}$ using LPIPS. This reward operates on deep feature representations, capturing structural layouts, object identities, and fine-grained texture patterns that are not fully reflected by photometric statistics:
\begin{equation}
r_{\mathrm{perc}} = \exp\Bigl(-\operatorname{LPIPS}(\mathbf{X}_e,\mathbf{X}_{gt})\Bigr),
\end{equation}
where a smaller perceptual distance yields a higher reward.
This reward provides a complementary supervision signal to $r_{\mathrm{photo}}$, encouraging the edited image to preserve the underlying scene structure. Altogether, these dual components ensure that the Photographic Artist achieves aesthetic enhancement while strictly forbidding unintended structural deviations. More importantly, with the design of disentangled and interpretable reward assignment, it remains sensitive to subtle retouching operations.

\paragraph{Is It Simply Reducing Distance to GT?}

Formally, yes — the optimization reduces the distance between the edited image and its ground-truth image in the chosen feature spaces. However, this is fundamentally different from pixel-wise reconstruction, which would collapse diversity and over-penalize permissible tonal variations. Instead, $r_{\mathrm{photo}}$ constrains the alignment of low-level photometric properties (e.g., exposure, color balance, etc.), while $r_{\mathrm{perc}}$ enforces structural and semantic consistency.

\paragraph{Why Is Multi-Level Reward Effective?}
Most existing diffusion-based RL frameworks primarily emphasize semantic-level evaluation, which is well-suited for high-level editing tasks such as object insertion, removal, or stylistic transformation. However, automatic photographic enhancement is fundamentally a continuous aesthetic refinement problem, where improvements are often subtle, incremental, and dominated by photometric variations rather than discrete semantic changes.
To address this gap, we decompose the reward into three complementary levels: (i) Semantic Compliance, which captures \textit{what to do} by aligning with high-level editing instructions; (ii) Photometric Control, which regulates \textit{how} low-level attributes such as exposure, contrast, and color temperature \textit{should change} to obtain a visually pleasing image; and (iii) Perceptual Consistency, which ensures that structural layout, object identity, and scene coherence are preserved and thus \textit{what should not break} during the editing process.
This hierarchical reward formulation enables the model to jointly reason over instruction adherence, fine-grained attribute manipulation, and structural stability.

\subsection{Reward Design for Photographic-aware Understanding}
The Image Critic is responsible for image quality/aesthetic reasoning, scoring, and edit suggestion generation. To align its response with {automatic photographic enhancement}, we propose a structured reward design to capture both global aesthetic judgment and fine-grained improvement guidance.

\paragraph{Reasoning Format Reward}
To enforce structured aesthetic analysis, we define a Reasoning Format Reward that penalizes deviations from the canonical template [reasoning → suggestion → score], which measures template compliance for the Chain-of-Thought reasoning process.

\paragraph{Score Ranking Reward}
To align the scoring capability with the generative module, we construct input–edited pairs $(\mathbf{X}, \mathbf{X}_e)$, where $\mathbf{X}_e$ is produced by the Photographic Artist. 
The edited result $\mathbf{X}_e$ are re-fed into the Image Critic to obtain its quality score. We then construct a pairwise ranking signal that reflects whether the generated edit leads to an improvement over the input. This ranking signal is used to define the following aesthetic-aware reward:
\begin{equation}
r_{\mathrm{rank}} =
\begin{cases}
1, & \text{if } \mathcal{S}(\mathbf{X}) < \mathcal{S}(\mathbf{X}_e) \\
0, & \text{otherwise}
\end{cases}
\end{equation}
where $\mathcal{S}({\mathbf{X}})$ and $\mathcal{S}({\mathbf{X}}_e)$ denote the predicted quality score of the input image and edited image from the Image Critic, respectively. This reward provides a consistency signal between the input assessment and the resulting edited image. Specifically, for images with the same underlying content, the edited result is expected to be of higher quality if the applied restoration or retouching is effective. By re-evaluating the edited image, the Image Critic is encouraged to correctly recognize such improvements and assign higher scores accordingly. In this way, the model learns a consistent ordering in which quality-enhanced versions are ranked above their original counterparts.

\paragraph{Edit Suggestion Reward via Exploration}
To generate effective and actionable edit suggestions, we integrate three complementary data sources: 
(i) image restoration datasets with explicit degraded-restored pairs, 
(ii) color grading datasets (e.g., FiveK), and 
(iii) high-aesthetic images from AVA augmented with synthetic degradations.
Instead of relying on deterministic ground-truth edit annotations, 
we acknowledge that photographic retouching inherently lies in a one-to-many solution space, 
where multiple valid enhancement strategies may exist for a given image. 
Therefore, we encourage the Image Critic to actively explore plausible edit suggestions.

For image restoration datasets, where degradations are well-defined, we impose explicit semantic constraints on the predicted suggestions. 
Specifically, the model is required to produce corresponding corrective operations (e.g., \textit{remove haze} for foggy images, \textit{deblur} for motion blur), ensuring basic restoration capability and preventing semantic mismatch.
For retouching data, including FiveK pairs and AVA-based synthesized pairs, we adopt an exploration-driven supervision strategy. 
Given an input image $\mathbf{X}$ and its higher-aesthetic counterpart $\mathbf{X}_{gt}$, the Image Critic first predicts a set of color- and tone-related edit suggestions $\mathcal{E}$. 
These suggested operations are then applied to $\mathbf{X}$ in a rule-based manner to produce a pseudo-edited reference image $\mathbf{X}_{ref}$, which approximates the effect of executing the inferred edits. 
We then define the reward by evaluating whether $\mathbf{X}_{ref}$ moves closer to $\mathbf{X}_{gt}$ in photometric attribute space, while simultaneously deviating from the original input $\mathbf{X}$. Similar to the previously introduced $r_{photo}$, we measure attribute-level distances and quantify the degree to which the suggested edits reduce the gap to the target image, which encourages the model to learn effective editing suggestions that yield improvements towards the ground-truth target.

This design enables robust and flexible edit suggestion reasoning, which is crucial for handling real-world photographic enhancement scenarios.
By jointly leveraging explicit supervision from image restoration tasks and exploration-based learning from retouching data, 
the Image Critic learns to produce interpretable and actionable edit suggestions, 
bridging the gap between visual understanding and controllable image enhancement.

\section{Experiments}
\subsection{Dataset}
\paragraph{Datasets for Image Critic.}  The Image Critic is trained to perform fine-grained aesthetic reasoning and distortion-aware analysis, score the image quality and provide edit suggestions. To this end, we leverage annotated IQA datasets and multiple task-specific image distortion datasets. (1) IQA datasets, including KonIQ-10K \cite{koniq}, SPAQ \cite{spaq}, and KADID-10K \cite{kadid}, all provide mean opinion score (MOS) scores, with KADID-10K offering distortion-aware annotations; (2) Image Distortion Datasets, including FoundIR \cite{foundir} (for deblurring, dehazing, low-light enhancement tasks), RealBlur \cite{realblur} (for deblurring task), TMM22 \cite{moire} (for moire removal task), LOL \cite{lol} and LOL-v2 \cite{lolv2} (for low-light enhancement task), ISTD~\cite{istd}, RDD~\cite{rdd} and SRD~\cite{srd} (for shadow removal task). These datasets provide paired degraded-restored images, allowing us to synthetically construct reasoning annotations corresponding to distortion types. 

\paragraph{Datasets for Photographic Artist.} The Photographic Artist is trained for multi-attribute image restoration and aesthetic retouching. (1) Image Restoration Datasets: We use the same restoration datasets for supervised generation training, optimizing flow-matching objectives between degraded and restored images; (2) Custom Retouching Dataset: To support fine-grained aesthetic editing, we construct a retouching dataset covering exposure adjustment, contrast enhancement, saturation modulation, correlated color temperature (CCT) and depth-of-field (bokeh-like effect) manipulation. For color- and tone-related adjustment, we adopt the FilmSet dataset \cite{filmset} to synthesize color grading pairs by applying attribute-specific perturbations to base images, simulating varying degrees of changes. For depth-of-field editing, RealBokeh \cite{realbokeh} and BokehDiff \cite{bokehdiff} are selected to construct image pairs with varying blur intensities.

\paragraph{Datasets for Unified Understanding and Generation.} We further utilize image pairs from FiveK dataset~\cite{fivek}, and high-aesthetic images from AVA dataset \cite{ava}. For the AVA subset, random synthetic degradations are applied to construct degraded–GT pairs with corresponding edit operations.

\paragraph{Dataset Statistics.}
In total, during foundation pretraining (Stage I), the Image Critic is trained on approximately 80K annotated samples, while the Photographic Artist is trained on approximately 160K paired images with corresponding instruction prompts. For unified understanding and generation, we utilize around 30K samples in Stage II and 18K samples in Stage III. Overall, our training pipeline leverages a diverse corpus spanning image quality assessment, restoration, and aesthetic enhancement tasks, covering both annotated and paired data with varied distortion types and aesthetic attributes.

\subsection{Implementation Details}
{We implement a three-stage training pipeline to optimize the Image Critic and the Photographic Artist. In Stage I (Foundation Pre‑training), we separately perform supervised fine‑tuning (SFT) on each model using the AdamW optimizer with a learning rate of $1\times 10^{-5}$. In Stage II (Reasoning‑Conditioned Adaptation), we continue SFT but update only the Photographic Artist with a learning rate of $1\times 10^{-5}$, to adapt it to reasoning‑guided editing. In Stage III (Coordinated Reasoning‑to‑Generation Reinforcement Learning), we jointly train both modules using two separate AdamW optimizers within the same training loop, with learning rates of $3 \times 10^{-4}$ for the Photographic Artist and $1 \times 10^{-5}$ for the Image Critic, respectively. All experiments are conducted on 8 NVIDIA A100 GPUs.}

\newcommand{\best}[1]{\textbf{#1}}
\newcommand{\second}[1]{\underline{#1}}

\subsection{Main Results}

\begin{table*}[t]
\centering
\caption{Comparison on Automatic Photographic Enhancement. The best results are highlighted in \best{bold}, and the second-best results in \second{underline}.}
\label{tab:reasoning_comparison}
\resizebox{0.7\textwidth}{!}{
\begin{tabular}{l cccccc}
\toprule
Method & MUSIQ$\uparrow$ & NIMA$\uparrow$& DINO$\uparrow$ &CLIP$\uparrow$&FID$\downarrow$ &LPIPS$\downarrow$ 
\\
\midrule
% \cellcolor{gray!20} \textit{Instruction-based models } \\
Instruct-Pix2Pix \cite{InstructPix2Pix} 
 &60.48 &5.27&0.20&0.45&224.90 &0.68
\\
FLUX2.Dev \cite{flux2dev} 
&\best{72.94}&\second{5.67}&0.93&0.88&76.26&0.26
\\
Qwen-Image-Edit \cite{qwenimage}  
&{68.63}&5.57&\second{0.96}&\second{0.93}&\second{42.81} &\second{0.17}
\\
OmniGen2 \cite{omnigen2} 
&67.52 &\best{5.69}&0.81&0.85&92.78 &0.41
\\
Step1X-Edit \cite{Step1X-Edit} 
&66.34 &5.31&0.95&0.89&48.48 &0.19

\\
\hline
\rowcolor{purple!5} Ours 
&\second{69.52}&5.66 &\best{0.98}&\best{0.96}&\best{27.96} &\best{0.10}\\
\bottomrule
\end{tabular}
}
\end{table*}

\begin{table*}[t]
\centering
\caption{Comparison on multi-edit instruction adherence where restoration and retouching tasks are combined. The best results are highlighted in \best{bold}, and the second-best results in \second{underline}.}
\label{tab:instruction_comparison}
\resizebox{0.7\textwidth}{!}{
\begin{tabular}{l cccccc}
\toprule
Method & PSNR$\uparrow$ & SSIM$\uparrow$ & LPIPS$\downarrow$ & FID$\downarrow$ &DINO$\uparrow$ &CLIP$\uparrow$\\
\midrule
Instruct-Pix2Pix \cite{InstructPix2Pix} 
&13.84 &0.49 &0.52 &186.41 &0.49 &0.60\\
FLUX2.Dev \cite{flux2dev} 
&15.93 &\second{0.71} &\second{0.20} &46.70 &\second{0.96} &\second{0.90}\\
Qwen-Image-Edit \cite{qwenimage}  
&14.76 &0.44 &0.24 &52.89 &0.93 &\second{0.90}\\
OmniGen2 \cite{omnigen2}   
&12.56 &0.33 &0.47 &97.36 &0.82 &0.82 \\
Step1X-Edit \cite{Step1X-Edit}  
&\underline{17.05} &0.62 &0.21 &\second{38.91} &0.94 &0.88\\
\hline
\rowcolor{purple!5} Ours  
&\best{21.05} &\best{0.82} &\best{0.09} &\best{22.93} &\best{0.97} &\best{0.96}    \\
\bottomrule
\end{tabular}
}
\end{table*}

\subsubsection{Automatic Photographic Enhancement}
We evaluate the automatic photographic enhancement capability of {SmartPhotoCrafter} against several open-source generative baselines \cite{omnigen2,Step1X-Edit,flux2dev,qwenimage,InstructPix2Pix}. All methods are required to automatically identify image deficiencies and generate an enhanced version. This setup reflects a realistic scenario where models must autonomously infer appropriate global and local adjustments.
For evaluation, we construct a comprehensive test suite covering multiple enhancement scenarios, including the FiveK dataset for general photographic retouching, a degraded subset derived from the AVA dataset containing high-aesthetic but synthetically corrupted images, and image restoration testsets to assess robustness under low-level degradations. This combination enables a holistic evaluation across standard retouching, aesthetic restoration, and degradation-aware enhancement settings.
Given the inherently subjective nature of photographic enhancement, where multiple visually pleasing outputs may coexist, we adopt a suite of metrics for comprehensive evaluation: (1) perceptual quality using MUSIQ~\cite{ke2021musiq} and NIMA~\cite{nima}, and (2) semantic and structural consistency with ground-truth images using DINO~\cite{dino}, CLIP~\cite{clip1,clip2}, FID~\cite{fid}, and LPIPS~\cite{lpips}. The ground-truth represents one plausible sample from a distribution of valid photorealistic enhancements, and these reference-based metrics measure high-level semantic similarity rather than strict pixel-wise correspondence.

Quantitative results are reported in Table~\ref{tab:reasoning_comparison}. 
{For reference, the ground-truth images achieve a MUSIQ score of 70.96 and a NIMA score of 5.23.}
The proposed method achieves competitive perceptual quality, ranking second on MUSIQ while maintaining strong performance on NIMA. Besides, it outperforms all baselines in semantic alignment and distribution fidelity, achieving the best scores on both DINO and CLIP, as well as the lowest FID and LPIPS. 
It can be seen that FLUX2.Dev~\cite{flux2dev} achieves strong performance on MUSIQ, indicating high perceived image quality. However, its lower performance on FID and structural alignment metrics suggests a tendency to deviate from the photorealistic distribution and introduce perceptual shifts. In contrast, our method achieves a more balanced improvement across both perceptual quality and distributional consistency, demonstrating better photorealistic editing while enhancing overall aesthetic quality.
These results indicate that SmartPhotoCrafter effectively enhances visual quality while preserving high-level semantics, and ensuring distributional consistency with photographic images.

\begin{figure}[t]
    \centering
    \includegraphics[page=4, width=\linewidth]{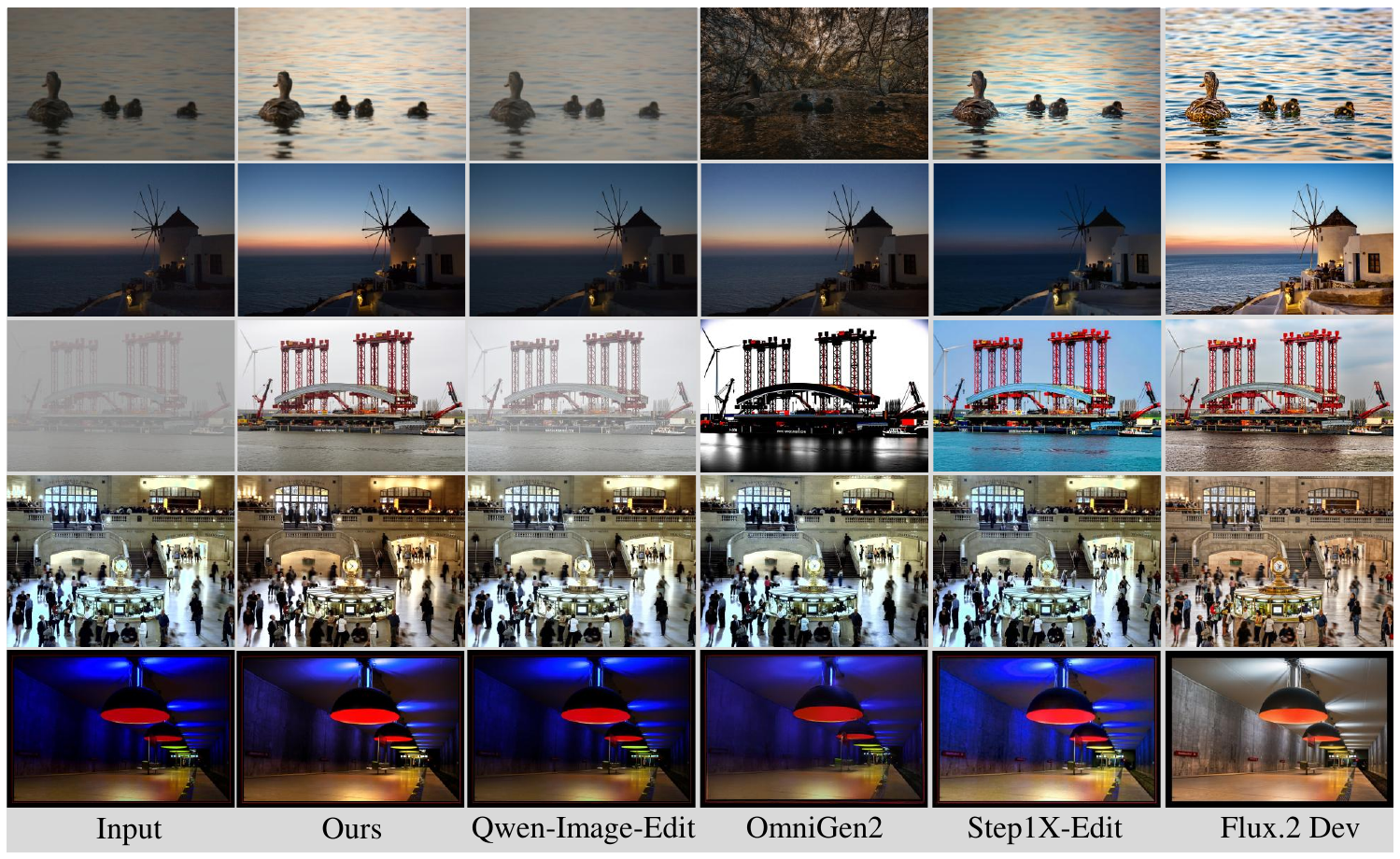}
    \caption{Visual comparison of automatic photographic enhancement task across different methods.} \label{vis_multi_auto}
\end{figure}

\begin{figure}[t]
    \centering
    \includegraphics[page=5, width=\linewidth]{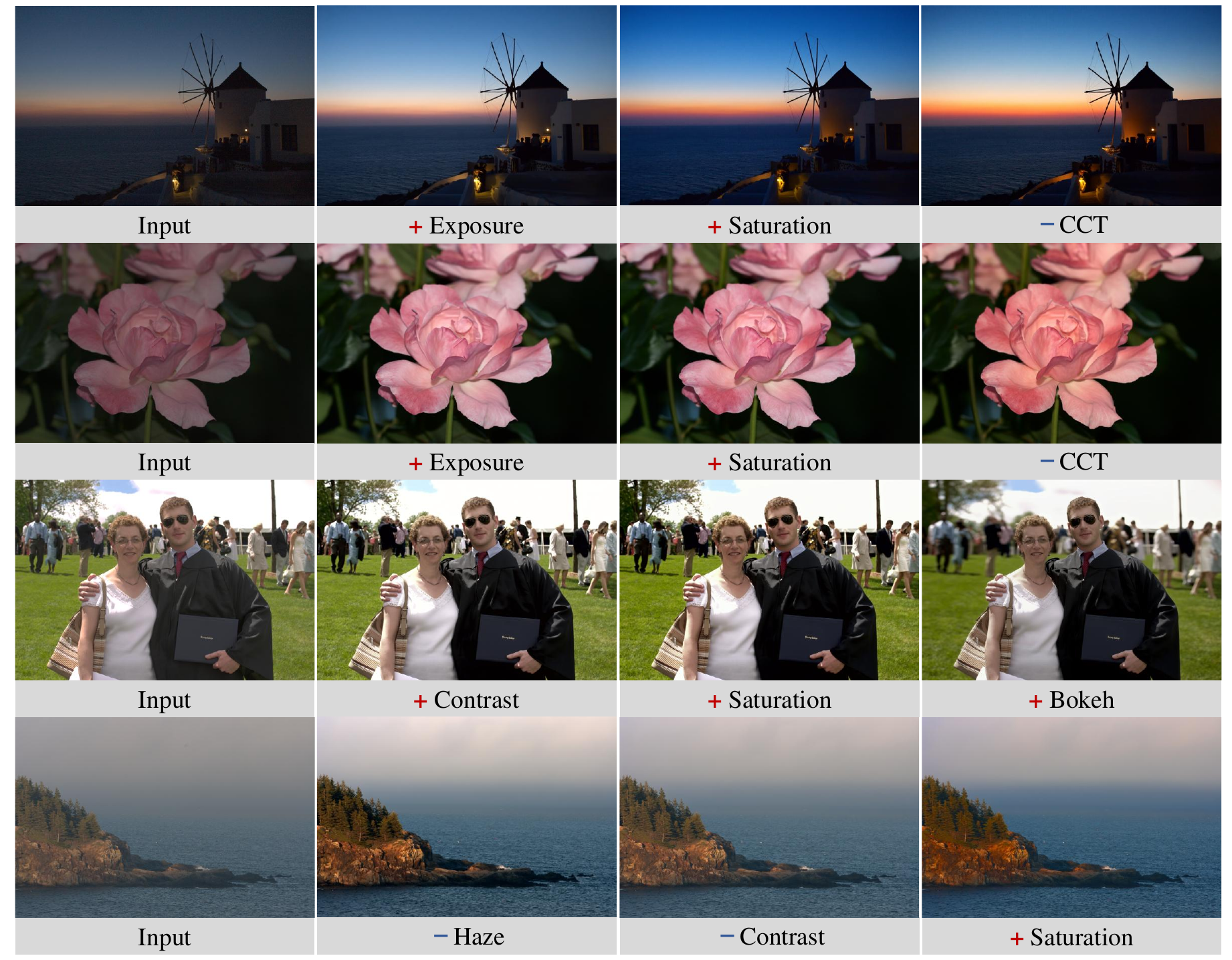}
    \caption{Examples showcase the cross-attribute instruction-based enhancement power, illustrating the instruction-following and generalization ability of SmartPhotoCrafter.} \label{vis_multi}
\end{figure}

\subsubsection{Multi-Edit Instruction Adherence}
We evaluate instruction-following capability of SmartPhotoCrafter in a multi-edit setting, focusing on whether the model can accurately execute a combination of restoration and retouching edits with explicit instructions. We apply random retouching operations on images with global degradations (blur and haze), and the model is expected to follow  combined restored and retouched instructions (e.g., remove blur, slightly decrease exposure and moderately increase saturation). We measure the performance using multiple reference-based perceptual and semantic metrics, including PSNR, SSIM, LPIPS, DISTS, FID, DINO and CLIP.

Quantitative results are summarized in Table~\ref{tab:instruction_comparison}. SmartPhotoCrafter outperforms the compared methods across all evaluation metrics, demonstrating superiority in multi-edit instruction adherence. In particular, it achieves the lowest LPIPS (0.09) and the highest DINO (0.97) and CLIP (0.96) scores, effectively reducing perceptual discrepancies and maintaining global consistency, even when applying complex, compositional modifications.
Meanwhile, SmartPhotoCrafter attains the highest PSNR and SSIM, reflecting strong fidelity to the reference images in both pixel-level reconstruction and structural consistency. The model also achieves the lowest FID score, suggesting that the generated results remain well-aligned with the distribution of high-quality reference images.

Overall, these results demonstrate that SmartPhotoCrafter can effectively perform combined restoration and retouching tasks. The consistent improvements across perceptual, structural, and semantic metrics demonstrate the ability of SmartPhotoCrafter on multi-edit instruction following.

\begin{table*}[t]
\centering
\caption{Comparison across Image Restoration tasks. Each series shows five metrics. The best results are highlighted in \best{bold}, and the second-best results are shown in \second{underline}.}
\label{tab:IR_comparison}
\resizebox{\textwidth}{!}{
\begin{tabular}{l ccccc ccccc}
\toprule
\multirow{2}{*}{\textbf{Method}} & \multicolumn{5}{c}{Deblur} & \multicolumn{5}{c}{Dehaze} \\
\cmidrule(lr){2-6} \cmidrule(lr){7-11}
 & LPIPS$\downarrow$ & FID$\downarrow$ & DISTS$\downarrow$ & PSNR$\uparrow$ & SSIM$\uparrow$& LPIPS$\downarrow$ & FID$\downarrow$ & DISTS$\downarrow$ & PSNR$\uparrow$ & SSIM$\uparrow$\\
\midrule
FLUX2.Dev~\cite{flux2dev}       & 0.20 &53.38 &0.12 &17.18 &0.60 &\second{0.09} &\second{36.99}&\second{0.07} &\second{21.17} &\best{0.88} \\
Qwen-Image-Edit~\cite{qwenimage} &\second{0.11} &\second{33.71}&\second{0.07} &24.53 &0.73 &0.23 &56.92&0.15 &14.35 &0.38 \\
% ICEdit        & 0.00 & 0.00 & 0.00 & 0.00 & 0.00 & 0.00 & 0.00 & 0.00 & 0.00 & 0.00 \\
FoundIR~\cite{foundir} &0.16 &50.12&0.11 &\best{25.92} &\best{0.77} &0.13 &39.81&0.11 &17.73 &0.82\\
MoCE-IR~\cite{moce-ir} &0.21 &50.68&0.13 &21.41 &0.71 &0.16 &39.34&0.12 &15.12 &0.72 \\
AdaIR~\cite{adair} & 0.19 & 51.35& 0.12 & 22.83 & 0.74 & 0.19 & 44.28& 0.15 & 15.99 & 0.74 \\
\hline
\rowcolor{purple!5} Ours &\best{0.07} &\best{21.85}&\best{0.05} &\second{24.73} &\second{0.76
} &\best{0.05} &\best{17.23}&\best{0.05} &\best{24.08} &\second{0.87} \\
\bottomrule
\end{tabular}
}
\end{table*}

\begin{table*}[t]
\centering
\caption{Ablation study on the reward design of Photographic Artist. The best results are highlighted in \best{bold}, and the second-best results are shown in \second{underline}.}
\label{tab:rl_ablation}
\resizebox{0.7\textwidth}{!}{
\begin{tabular}{c c c  c c c c c}
\toprule
SFT & RL (w/o $r_{\mathrm{photo}}$) & RL (Full) 
& MUSIQ$\uparrow$ &NIMA$\uparrow$ &FID$\downarrow$ &DINO$\uparrow$ &CLIP$\uparrow$\\
\midrule
\checkmark & & &{67.82} &{5.57} &\underline{30.61} &\underline{0.97} &\underline{0.95}\\
\checkmark & \checkmark &  &\second{68.25} &\underline{5.58} &38.51 &\underline{0.97} &0.94 \\
\checkmark & \checkmark & \checkmark  &\best{69.52}&\best{5.66} &\best{27.96}&\best{0.98}&\best{0.96} \\
\bottomrule
\end{tabular}
}
\end{table*}

\subsubsection{Qualitative Results}
As shown in Figure \ref{vis_multi_auto}, on the task of automatic photographic enhancement, SmartPhotoCrafter consistently improves color performance while preserving scene structure and textural details, producing natural and aesthetically pleasing results without introducing artifacts or content distortion. In contrast, although Flux.2 Dev~\cite{flux2dev} can generate visually appealing colors, it exhibits stronger AI-stylized appearance that deviates significantly from the input images, thereby reducing fidelity to the original photographs. The compared methods are either under-enhanced, leaving residual haze and dull tones, or over-enhanced, resulting in excessive saturation and contrast, and sometimes even generating hallucinated textures or altering scene semantics. Overall, SmartPhotoCrafter achieves a balance between aesthetic improvement and content preservation, producing high-quality results that remain faithful to the original image. More qualitative results of automatic photographic enhancement can be referred to Appendix~\ref{appendix:case_studies}.

{
In Figure \ref{vis_multi}, we visualize the results of different methods on the task of multi-edit instruction-following editing.  SmartPhotoCrafter supports joint control over multiple parameters (e.g., exposure, saturation, contrast) of real images based on natural language user instructions. Across different scenes and contents, our method enables continuous, composable, and fine-grained adjustments of multiple attributes according to user instructions, while maintaining stylistic consistency and structural stability, avoiding common issues such as color casts, artifacts, or semantic tampering. These result demonstrate that SmartPhotoCrafter learns a well-behaved and disentangled control space that generalizes to diverse image content and editing intents, enabling precise instruction-driven photo enhancement that is faithful to the original image.
}

\subsubsection{Comparison on Image Restoration Tasks}

We further evaluate the restoration capability of SmartPhotoCrafter on two representative low-level image restoration tasks, including deblurring and dehazing. For image deblurring, we adopt the testing set from FoundIR-Blur~\cite{foundir}, Realblur~\cite{realblur} and GoPro~\cite{gopro} datasets, while the testing set from FoundIR-Haze~\cite{foundir} is adopted for the image dehazing task.
Given that image restoration involves both fidelity recovery and perceptual quality improvement, we adopt the following metrics for evaluation: (1) reconstruction fidelity (PSNR and SSIM), (2) perceptual similarity (LPIPS, DISTS and FID).

Quantitative results are reported in Table~\ref{tab:IR_comparison}. On both deblurring and dehazing tasks, the proposed method consistently achieves the best or second-best performance across all metrics. In particular, SmartPhotoCrafter obtains the lowest LPIPS, DISTS and FID scores, indicating that the restored images are both perceptually closer to the ground truth and better aligned with the distribution of high-quality natural images. Meanwhile, competitive PSNR and SSIM values demonstrate that our method maintains strong reconstruction fidelity.

Notably, compared to task-specific restoration methods, SmartPhotoCrafter exhibits a more favorable balance between enhancement strength and content preservation. This can be attributed to the design that leverages high-level visual reasoning to guide image generation, allowing the model to perform implicit degradation correction as part of a broader enhancement process. These results suggest that, beyond automatic photographic enhancement, SmartPhotoCrafter also generalizes well to classical image restoration scenarios.

\subsection{Ablation Study}
\subsubsection{Effectiveness of Retouching-Aware Reward Design}
We conduct ablation studies on the Photographic Artist to analyze two key aspects:
(i) the effectiveness of reinforcement learning (RL), and (ii) the role of the Photometric Contral Reward $r_{\mathrm{photo}}$ in improving photometric quality. Experiments are conducted on a combination of the FiveK dataset (Input-ExpertC pairs) and the degraded AVA-GT paired dataset, where input images exhibit suboptimal styles and the GT images serve as expert-level references.

As shown in Table~\ref{tab:rl_ablation}, comparing SFT-only and SFT$+$RL (w/o $r_{\mathrm{photo}}$), we observe that introducing RL improves perceptual quality (MUSIQ: 67.82 → 68.25) and slightly boosts aesthetic quality (NIMA: 5.57 → 5.58). However, this comes at the cost of degraded distributional fidelity, as reflected by a worse FID score (30.61 → 38.51), while semantic alignment (DINO/CLIP) remains unchanged. This suggests that RL without explicit photometric constraints tends to over-optimize high-level perceptual or aesthetic objectives, potentially leading to distribution drift and less realistic outputs.

By further incorporating $r_{\mathrm{photo}}$, the full RL model achieves more balanced and consistent improvements. Specifically, FID is significantly reduced (38.51 → 27.96), while NIMA is further improved (5.58 → 5.66). In addition, DINO and CLIP scores reach their highest values (0.98 and 0.96), indicating enhanced semantic and structural alignment. The overall improvement across distributional, aesthetic, and semantic metrics demonstrates a more robust optimization behavior.

These results confirm that $r_{\mathrm{photo}}$ contributes to improved performance by introducing fine-grained photometric supervision. It effectively guides the model toward realistic tonal adjustments while preventing over-optimization. As a result, the combination of full rewards enables the model to achieve both high perceptual quality and faithful distribution alignment, leading to consistently superior and visually realistic editing performance.

\section{Conclusion}
In this work, we present SmartPhotoCrafter, a unified framework for automatic photographic image editing that integrates multimodal understanding with photorealistic image enhancement. By leveraging a vision-language model to infer detailed analysis to condition image synthesis, the proposed method bridges high-level aesthetic reasoning and multi-level visual transformation. During the last stage of model training, we design a set of rewards that jointly encourage image fidelity, consistency with inferred edits, and perceptual quality improvement. Experimental results demonstrate that SmartPhotoCrafter achieves strong performance across diverse enhancement scenarios, while maintaining naturalness and visual coherence. Overall, this work highlights the potential of coupling semantic understanding with generative modeling for intelligent photographic enhancement systems.
Nevertheless, the current framework primarily focuses on restoration and photometric adjustments, while preserving the original image content and structural layout. As a result, higher-level compositional factors remain unexplored.
Future work will investigate composition-aware enhancement beyond low-level adjustments. In addition, we will explore more tightly coupled optimization strategies that enable deeper interaction between semantic reasoning and image generation, allowing the Image Critic and Photographic Artist to mutually reinforce each other in a unified framework.

\bibliographystyle{splncs04}
\bibliography{main}

\newpage

\appendix
\section{More Qualitative Results}
\label{appendix:case_studies}

\begin{figure}[h]
\centering
\setlength{\tabcolsep}{2pt} % 控制列间距

\begin{tabular}{cccc}
\includegraphics[width=0.23\linewidth]{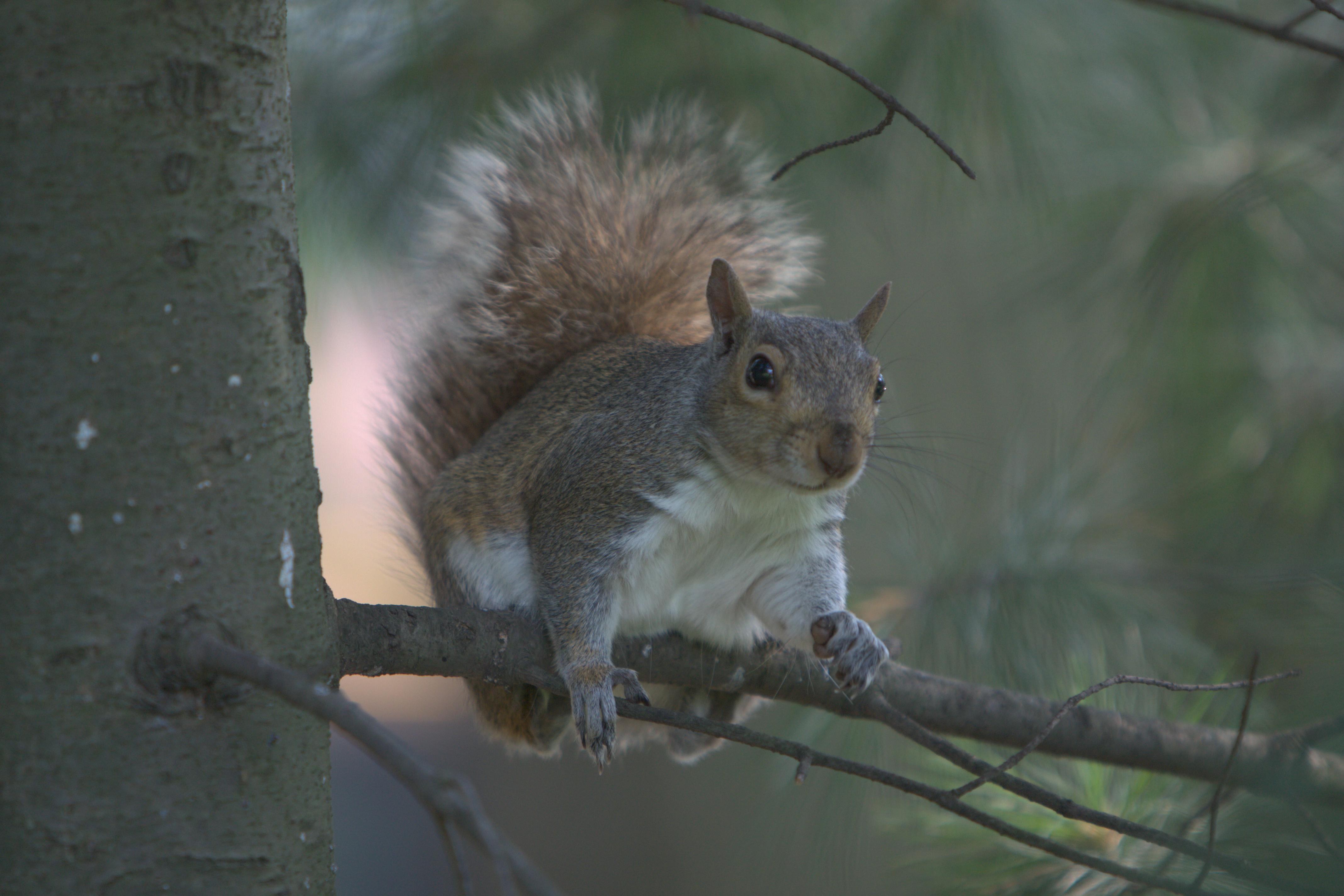} &
\includegraphics[width=0.23\linewidth]{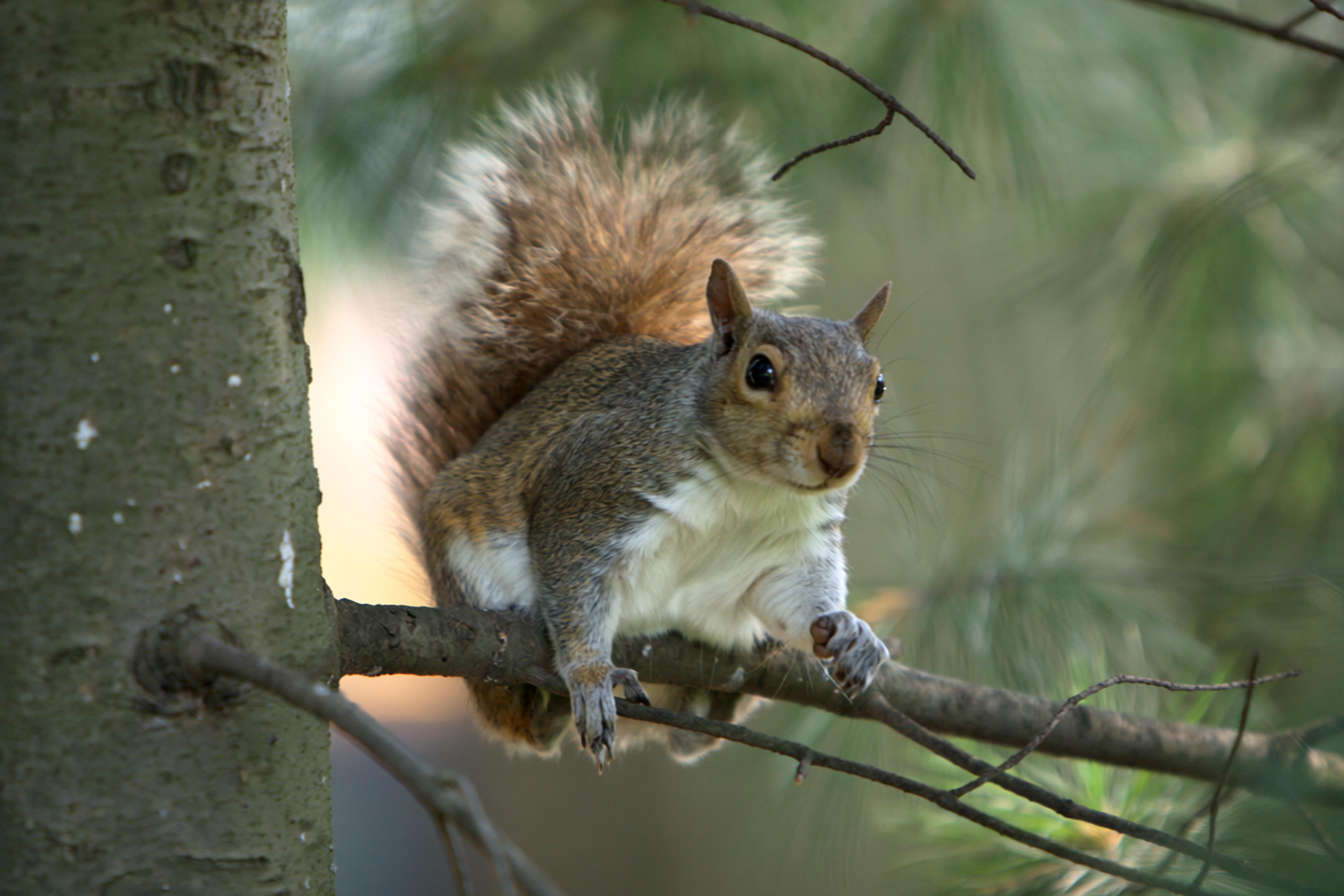} &
\includegraphics[width=0.23\linewidth]{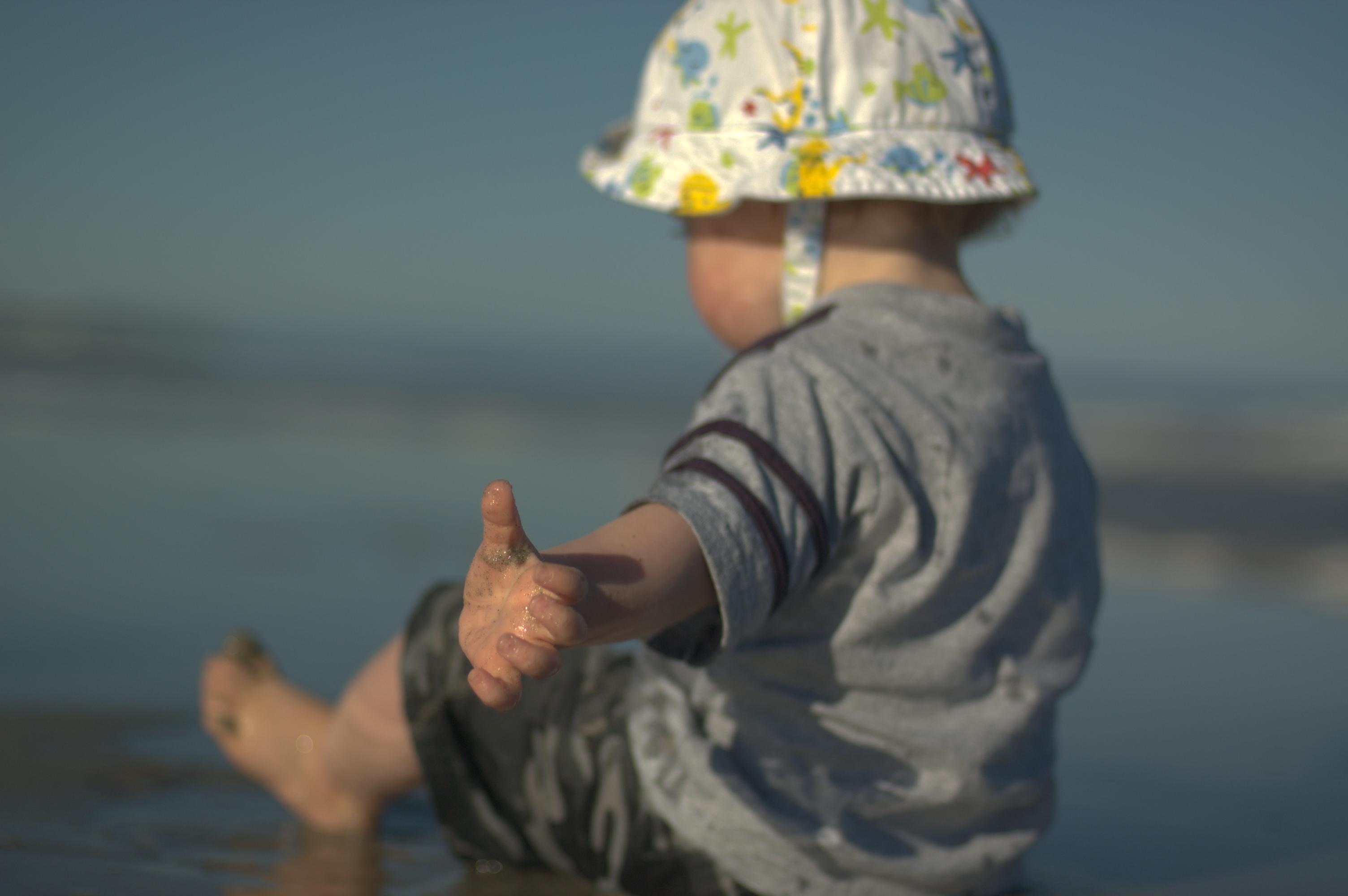} &
\includegraphics[width=0.23\linewidth]{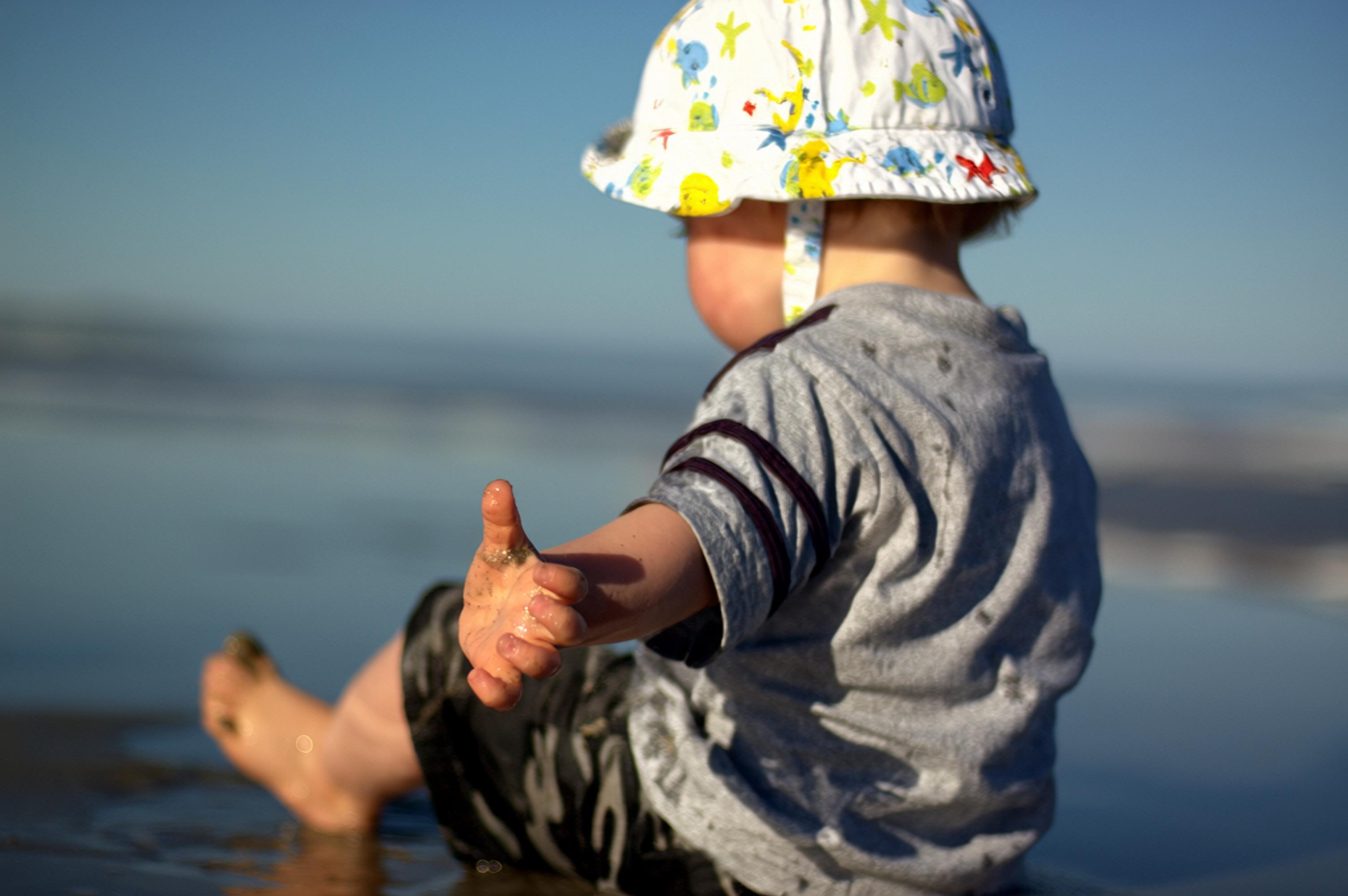} \\

\includegraphics[width=0.23\linewidth]{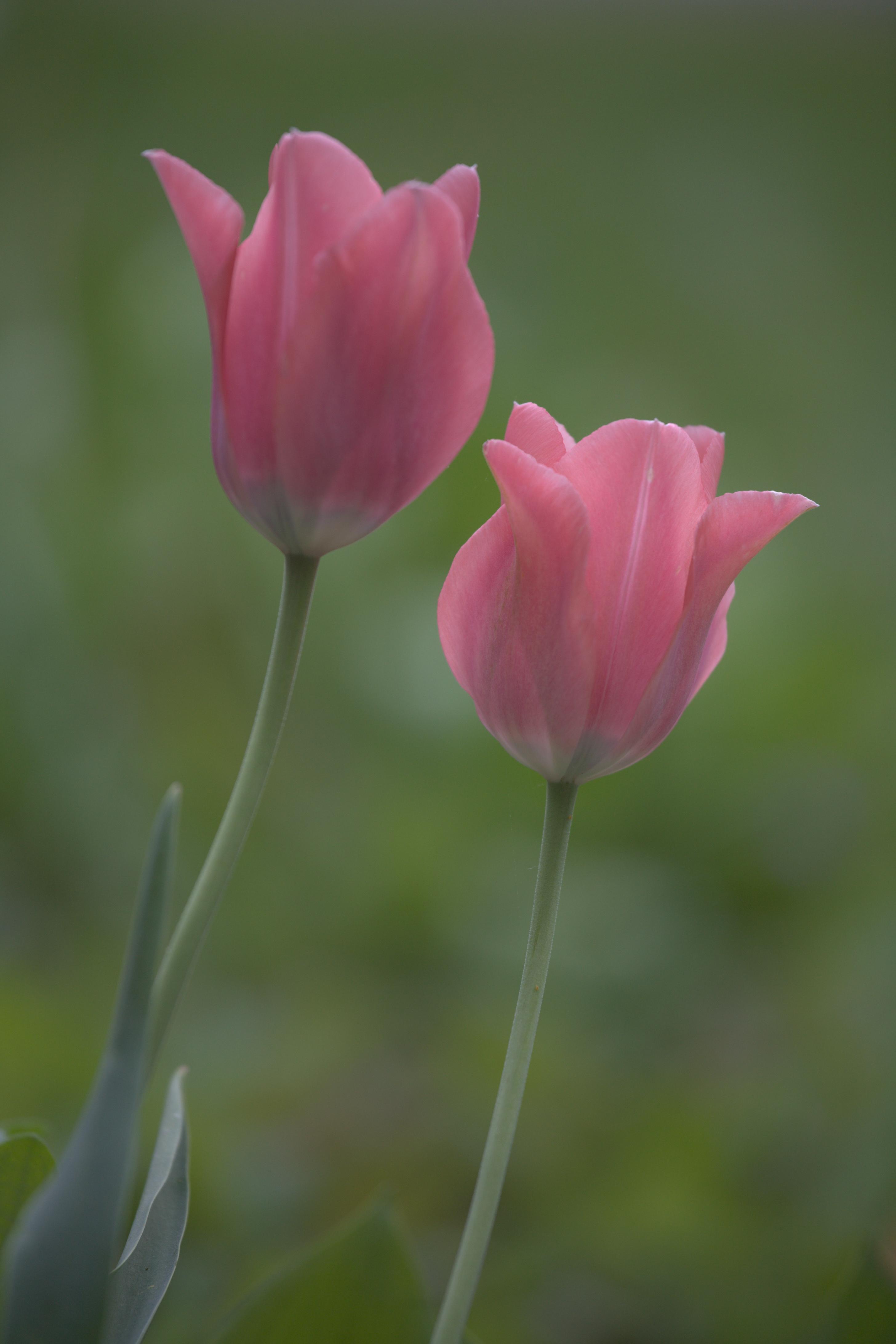} &
\includegraphics[width=0.23\linewidth]{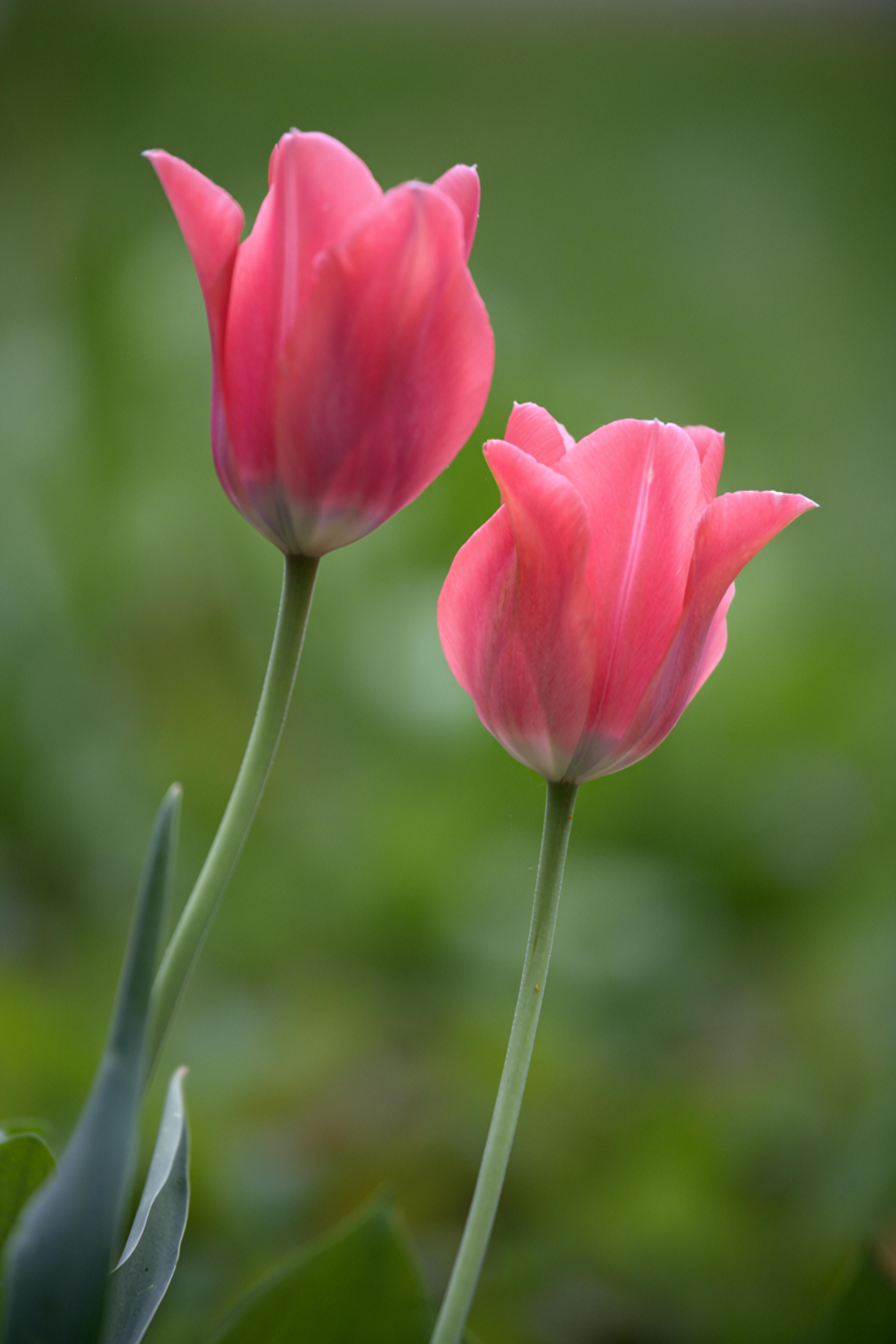} &
\includegraphics[width=0.23\linewidth]{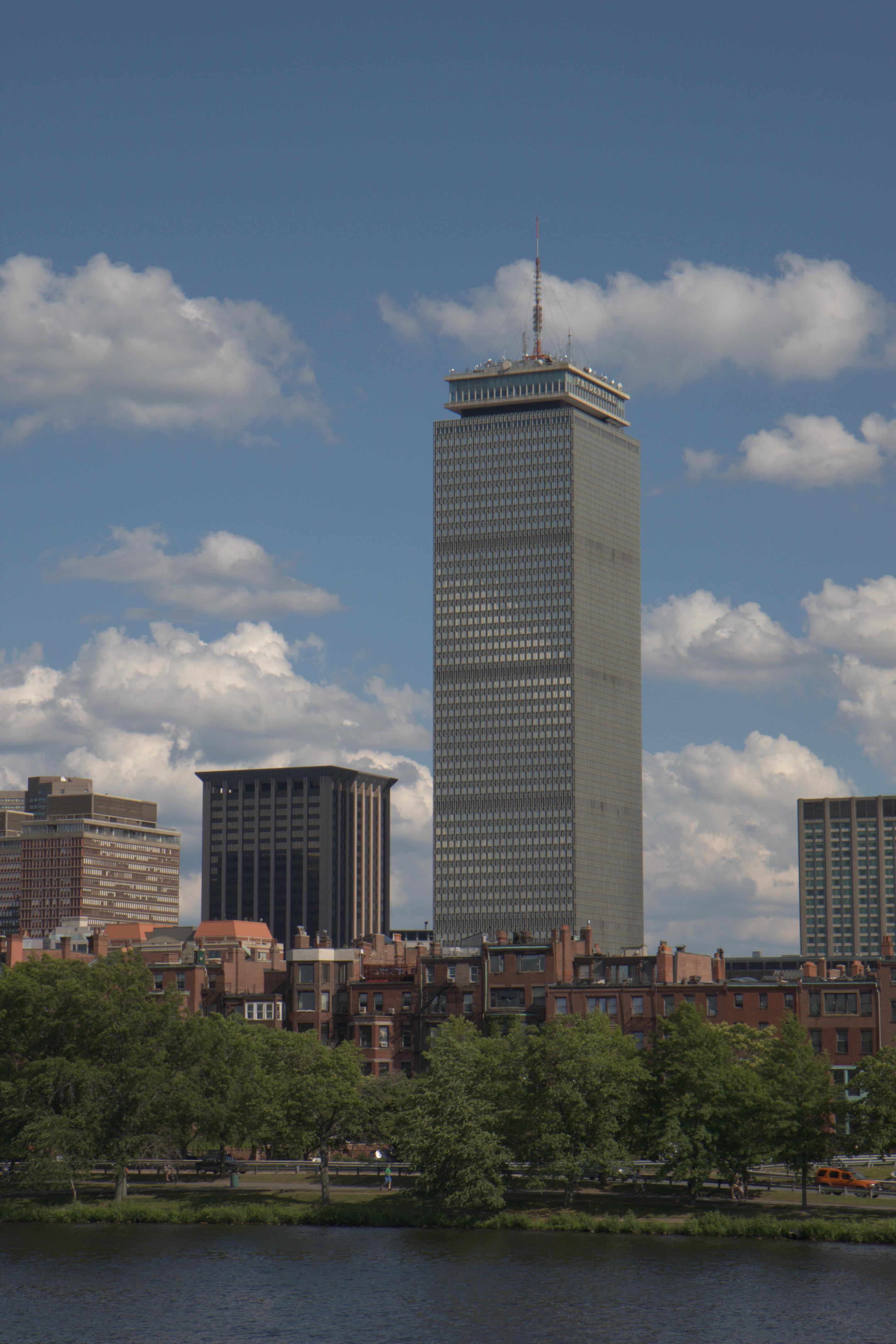} &
\includegraphics[width=0.23\linewidth]{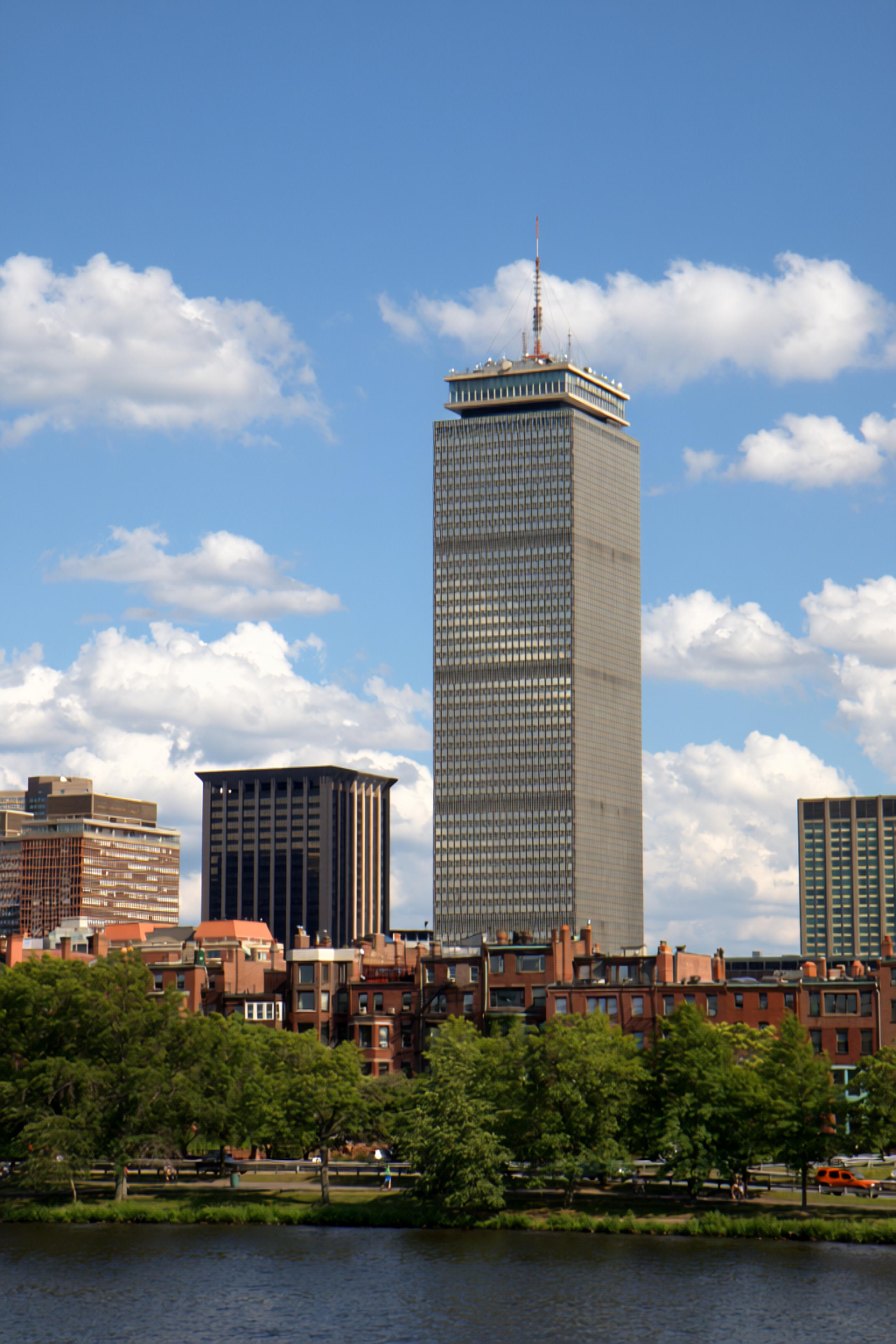} \\

\includegraphics[width=0.23\linewidth]{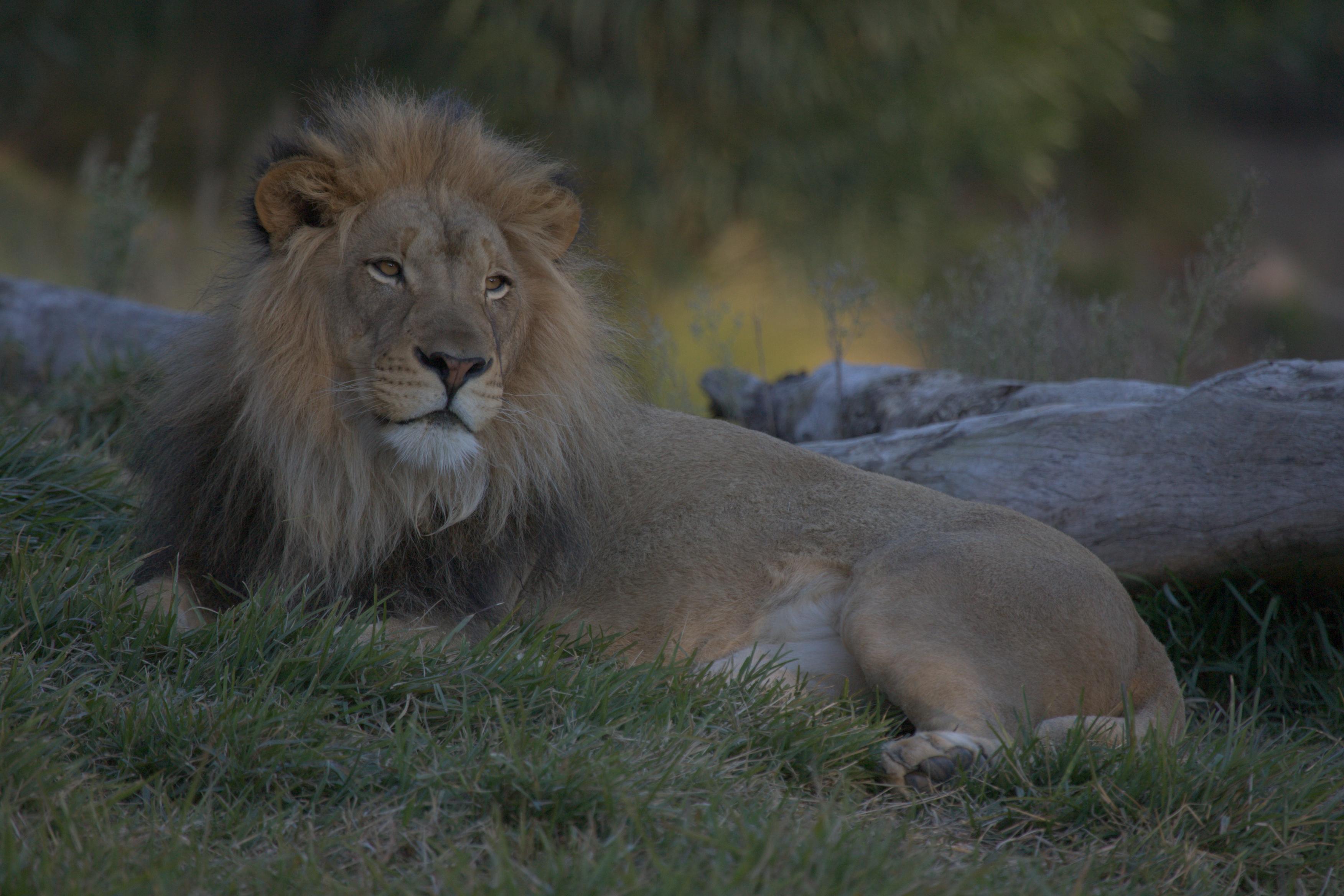} &
\includegraphics[width=0.23\linewidth]{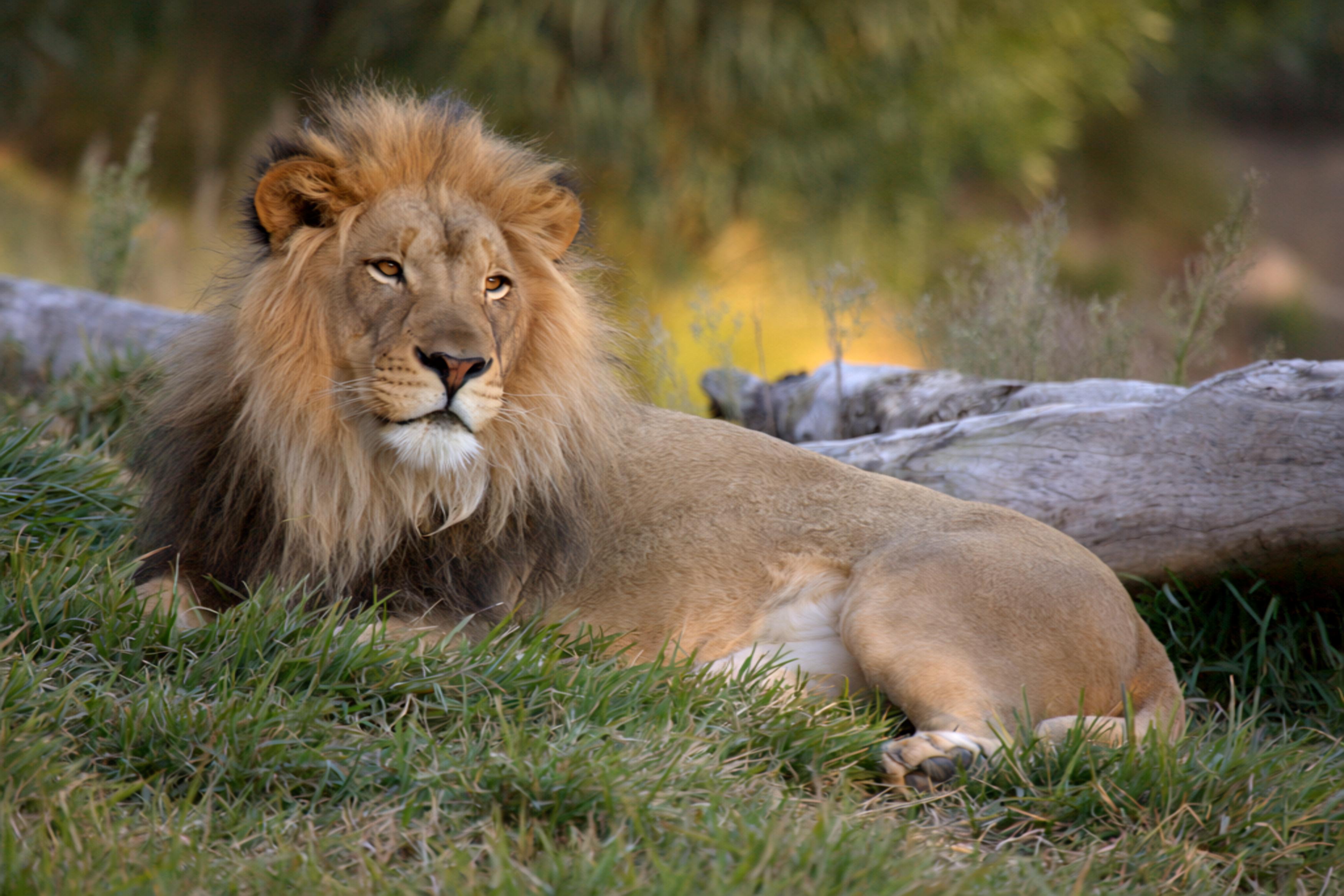} &
\includegraphics[width=0.23\linewidth]{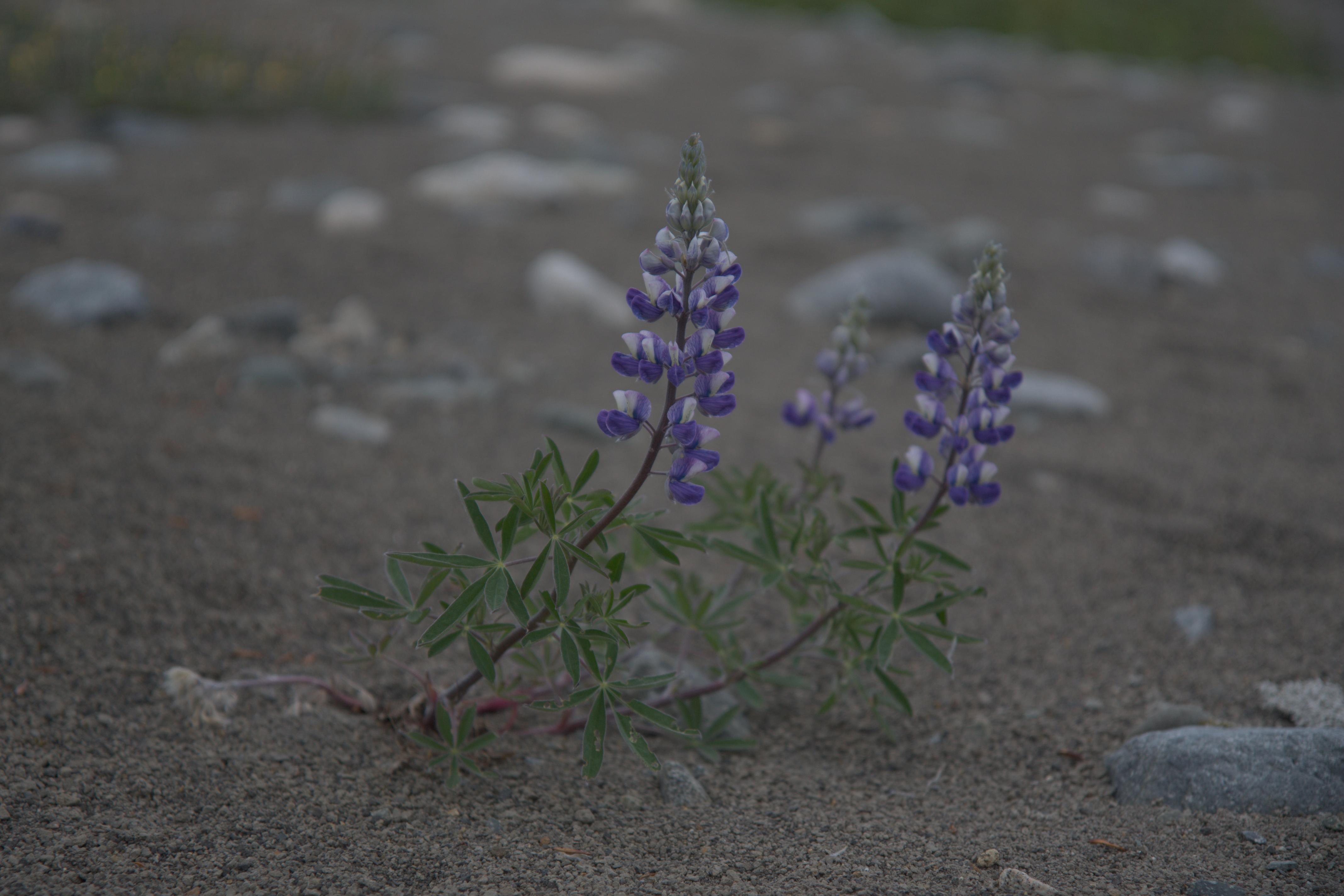} &
\includegraphics[width=0.23\linewidth]{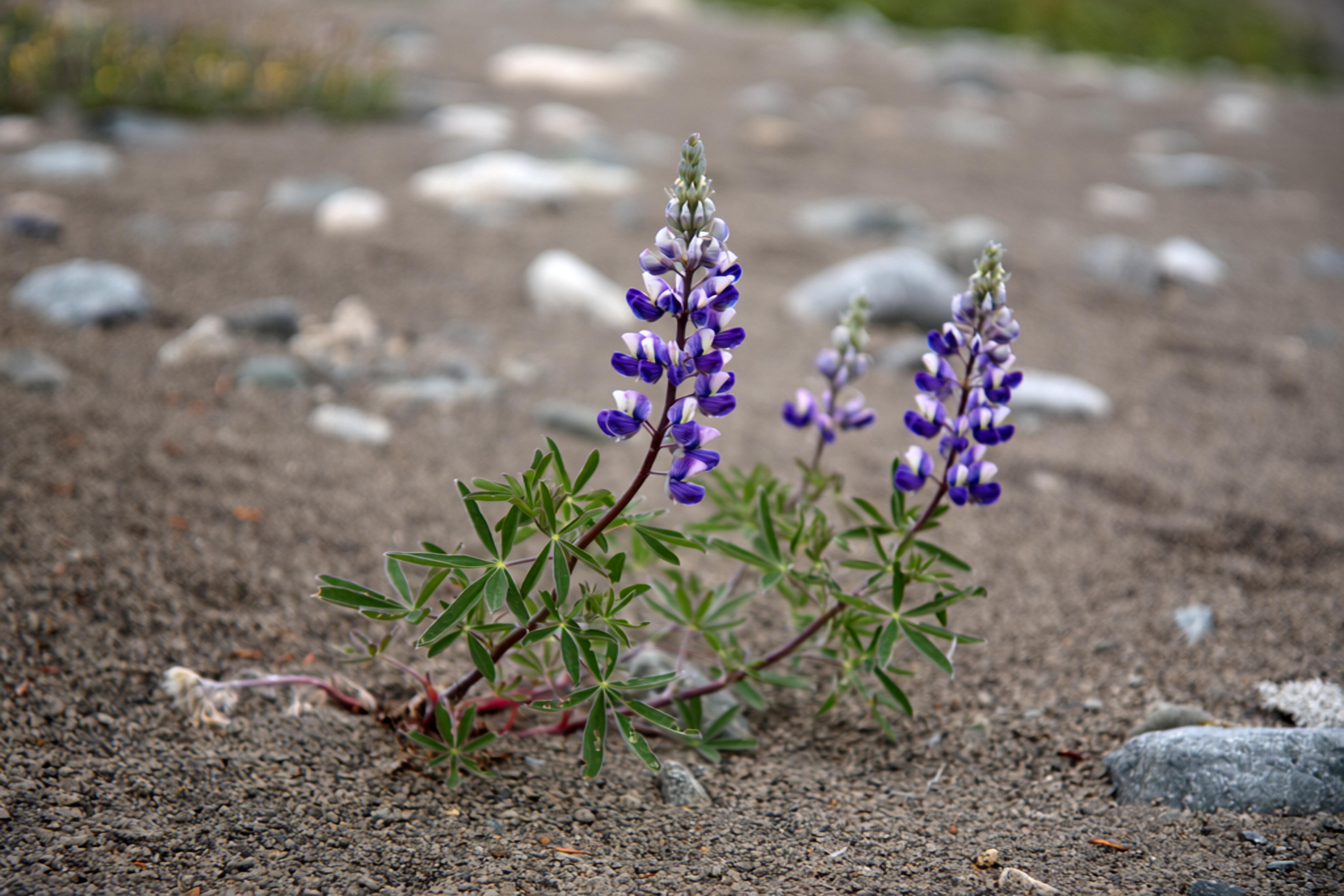} \\

\end{tabular}
\caption{More qualitative results on Automatic Photographic Enhancement.}
\label{fig:qualitative_results}
\end{figure}

We present additional qualitative examples of automatic photographic enhancement. As shown in Figure~\ref{fig:qualitative_results}, SmartPhotoCrafter produces visually pleasing results with improved aesthetics, better tonal balance, and enhanced details under diverse shooting scenarios. In particular, the enhanced images exhibit more appealing and well-balanced colors while faithfully preserving the original content and structural composition.
Importantly, SmartPhotoCrafter performs realistic enhancement rather than re-synthesis. The outputs remain photorealistic and consistent with the natural image distribution, avoiding artifacts or unnatural appearances with over-processed or generative results. Instead of altering the scene content, SmartPhotoCrafter performs style-consistent photographic improvements, achieving a desirable balance between fidelity and enhancement quality.

We further provide comprehensive case studies on the IQA dataset, including the full analysis consisting of the Image Critic outputs, input images, and the corresponding enhanced results by the Photographic Artist. These examples span images of varying quality levels, offering a more thorough evaluation of the model behavior. In the visualization, correct analysis are highlighted in blue, while inaccurate or suboptimal analysis are marked in red.

Overall, the results demonstrate that the Image Critic is capable of effectively analyzing diverse image attributes and summarizing actionable editing suggestions. Guided by these analysis, the Photographic Artist is able to generate improved enhancement results.

\begin{figure}[h]
    \centering
    % --- 第一部分：横向并排的两张图片 ---
    \begin{subfigure}[b]{0.49\textwidth}
        \centering
        \includegraphics[width=\textwidth]{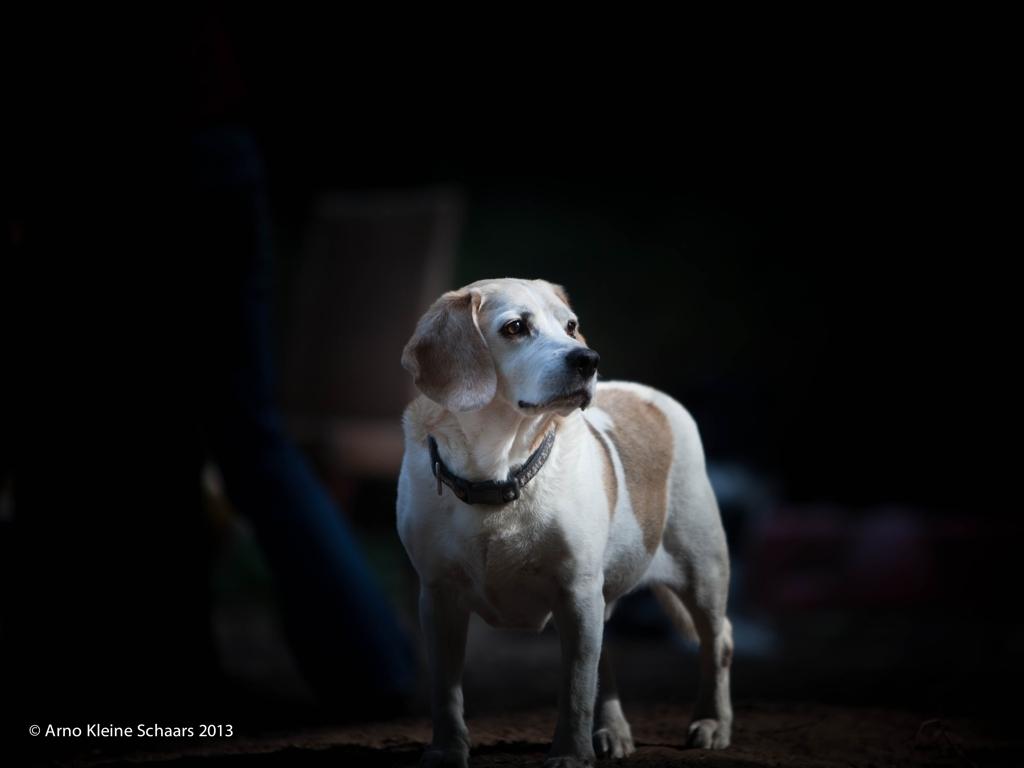} 
        \caption{Input Image}
    \end{subfigure}
    \hfill 
    \begin{subfigure}[b]{0.49\textwidth}
        \centering
        \includegraphics[width=\textwidth]{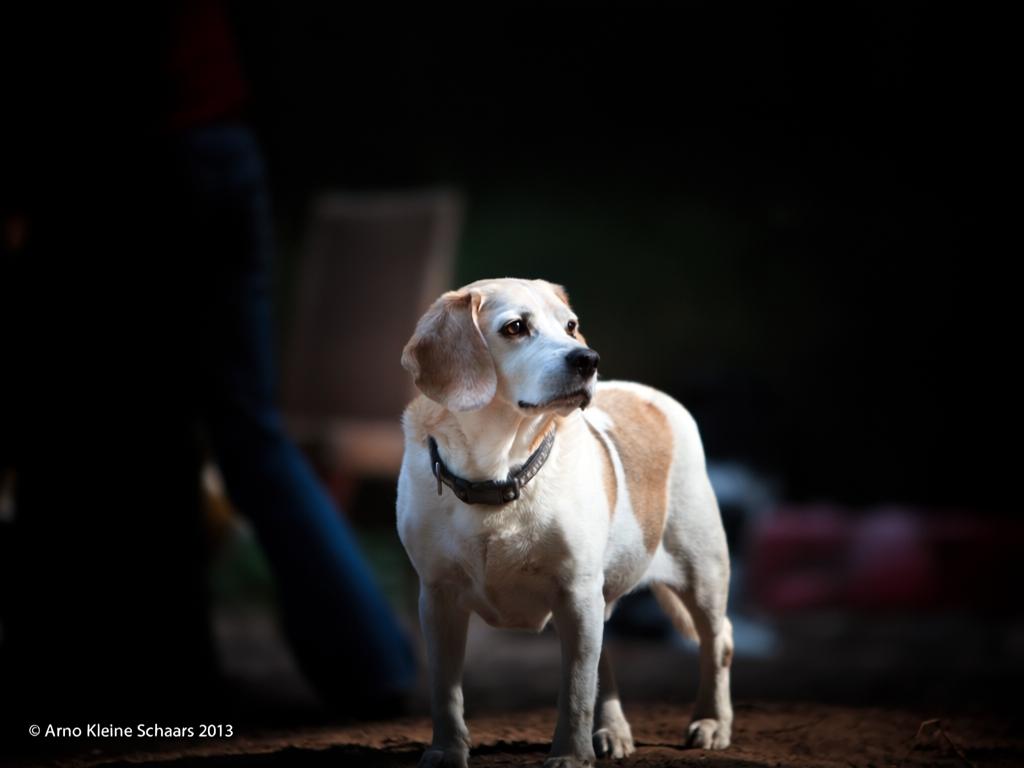} 
        \caption{Enhanced Image}
    \end{subfigure}
    
    % --- 第二部分：图片的总标题 (Case 1. GT Score) ---
    % \vspace{2mm}
    \caption{Ground-truth Score: 80.65}
    % {\small \textbf{Case 1. GT Score: 80.65}}
    \vspace{4mm}

    % --- 第三部分：推理结果表格 (无对比，标题居中) ---
    \small
    \begin{tabularx}{\textwidth}{X}
        \toprule
        % \multicolumn{1}{c} 确保标题在 X 列中居中
        \rowcolor{gray!10} \multicolumn{1}{c}{\textbf{Image Quality\&Aesthetic Reasoning}} \\ 
        \midrule
        1. Image Quality/Degradations Analysis: 
        The image exhibits \textcolor{blue}{no significant motion blur, defocus, or distortion}. However, there is noticeable \textcolor{blue}{low exposure, particularly in the darker areas} surrounding the dog. This creates a lack of detail in the background and could be considered a form of visual interference due to shadow occlusion. There is \textcolor{blue}{no apparent digital noise or compression artifacts} visible at this resolution.
        \\
        2. Color Performance\&Lighting Analysis:
        The \textcolor{blue}{saturation appears moderate}, with natural-looking colors. The \textcolor{blue}{white balance seems accurate}, as the dog's fur tones look true-to-life. However, the \textcolor{blue}{contrast is somewhat subdued due to the low exposure}, leading to less distinction between the subject and its surroundings. The lighting is focused on the dog, creating a dramatic effect, but it also results in some underexposed regions. \textcolor{blue}{Shadows are intentionally used to create depth}, but they obscure details in certain areas.
        \\
        3. Composition\&Layout Analysis:  
        The main subject, the dog, is clearly defined and positioned \textcolor{red}{slightly off-center, adhering to the rule of thirds}. The focus is sharp on the dog, drawing attention directly to it. The \textcolor{blue}{camera viewpoint is at eye level}, providing a natural perspective. The \textcolor{blue}{background blur is effective in isolating the subject}, though it might be slightly excessive, causing some loss of environmental context. The photography style leans towards a cinematic approach, emphasizing mood over clarity.
        \\
        
        4. Aesthetic Impression Analysis:
        The \textcolor{blue}{tone style is somewhat muted} due to the low exposure, giving the image a contemplative and serene feel. The \textcolor{blue}{creativity lies in the use of lighting and shadow to evoke emotion}. The emotional expression is \textcolor{blue}{calm and introspective}, with the dog appearing relaxed yet alert. The semantic richness is moderate, as the image tells a simple story of a quiet moment. The \textcolor{blue}{environment integrates well with the subject}, enhancing the overall mood. The context adaptability is good within the intended artistic style.
        \\
        
        5. Comprehensive Evaluation:
        The image has a \textcolor{blue}{strong aesthetic appeal due to its effective use of lighting and composition}. However, the \textcolor{blue}{low exposure affects the visibility of details} and could be improved. The color performance is generally good, but increasing the light and exposure would enhance the vibrancy. The composition is strong, with a clear focus on the subject, but \textcolor{blue}{reducing the background blur slightly could add more context}. The overall tone and emotional expression are compelling, making the image engaging despite the technical issue.  \\
        
        \midrule
        \rowcolor{gray!10} \multicolumn{1}{c}{\textbf{Edit Suggestions}} \\ 
        \midrule
        
        1. moderately increase light\&exposure; \\ 
        2. \textcolor{red}{slightly decrease the depth of field (enhance the bokeh effect)} \\

        \midrule
        \rowcolor{gray!10} \multicolumn{1}{c}{\textbf{Image Quality Scoring}} \\ 
        \midrule
        
        Predicted Score: 82/100\\

        \bottomrule
    \end{tabularx}
    \label{fig:case1}
\end{figure}

\begin{figure}[h]
    \centering
    % --- 第一部分：横向并排的两张图片 ---
    \begin{subfigure}[b]{0.49\textwidth}
        \centering
        \includegraphics[width=\textwidth]{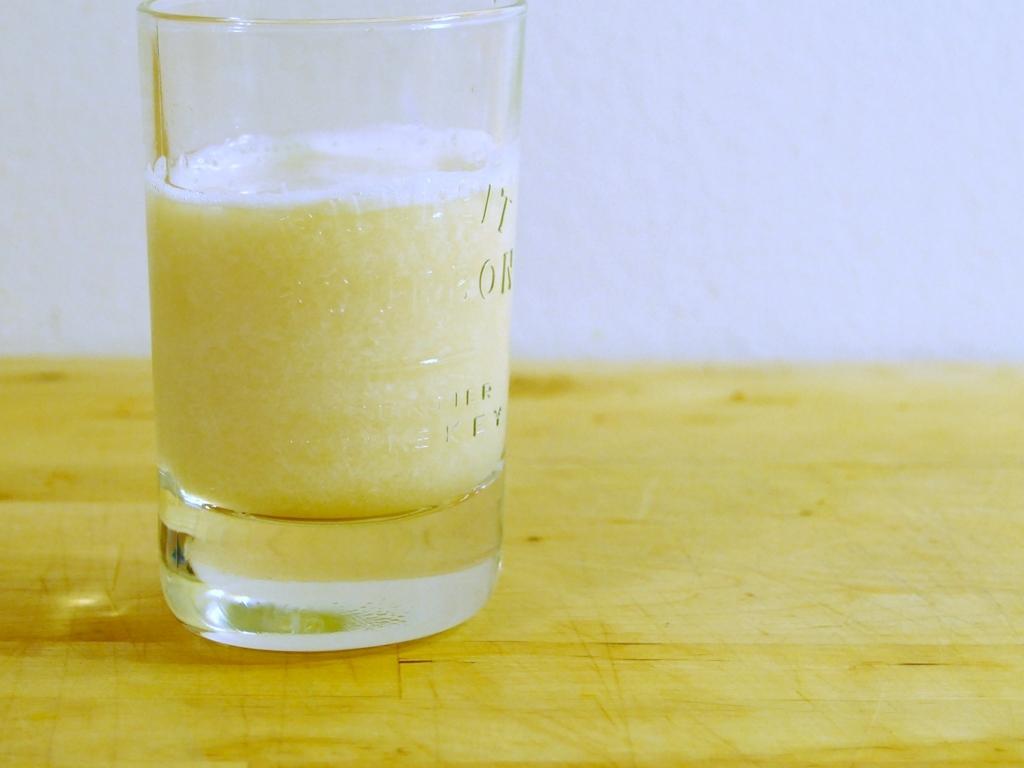} 
        \caption{Input Image}
    \end{subfigure}
    \hfill 
    \begin{subfigure}[b]{0.49\textwidth}
        \centering
        \includegraphics[width=\textwidth]{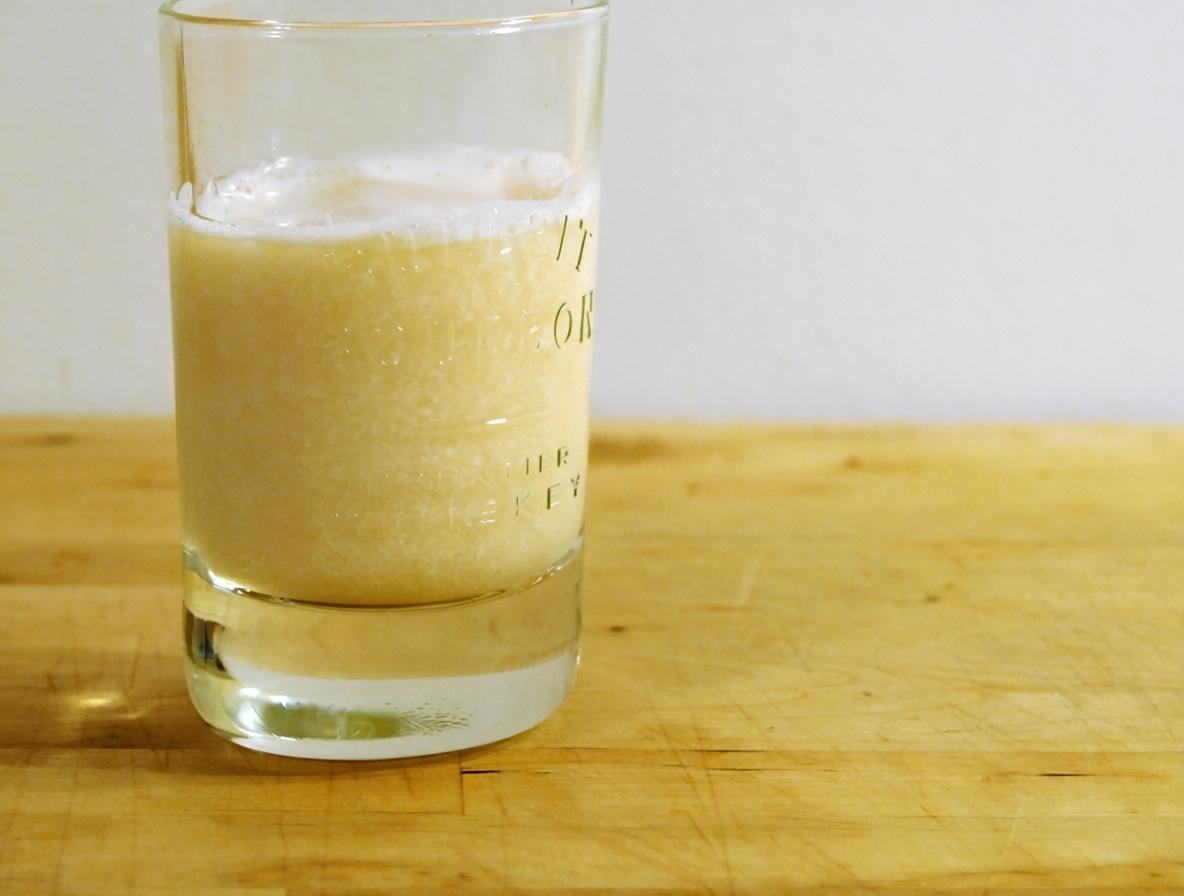} 
        \caption{Enhanced Image}
    \end{subfigure}
    
    \caption{Ground-truth Score: 66.19}
    \vspace{4mm}

    \small
    \begin{tabularx}{\textwidth}{X}
        \toprule
        % \multicolumn{1}{c} 确保标题在 X 列中居中
        \rowcolor{gray!10} \multicolumn{1}{c}{\textbf{Image Quality\&Aesthetic Reasoning}} \\ 
        \midrule
        1. Image Quality/Degradations Analysis: The image appears relatively \textcolor{blue}{free from significant distortions} such as motion blur, digital noise, or lens distortion. However, there is \textcolor{blue}{a slight overexposure} affecting the glass and its contents, causing some loss of detail in brighter areas. There is \textcolor{blue}{no noticeable glare, flare, or shadow occlusion} impacting the main subject. The distortion severity is moderate due to the overexposure.
        \\
        2. Color Performance\&Lighting Analysis:
        The \textcolor{blue}{saturation seems adequate}, providing a natural appearance to the drink and glass. The white balance appears slightly off, leaning towards a \textcolor{blue}{cooler tone, which might not accurately represent the warmth of the beverage}. The \textcolor{blue}{contrast is moderate but could be enhanced} for better differentiation between the drink and the background. The \textcolor{blue}{exposure issue} mentioned earlier affects the brightness levels, making parts of the image appear washed out.
        \\
        3. Composition\&Layout Analysis:  
        The main subject, the glass of drink, is \textcolor{blue}{centered and clearly visible}, adhering to standard composition rules. The \textcolor{blue}{focus is sharp on the glass}, ensuring it remains the primary point of interest. The \textcolor{blue}{camera viewpoint is straightforward}, capturing the subject at eye level. The \textcolor{blue}{depth of field is sufficient}, keeping both the foreground and background in focus. The photography style leans toward a \textcolor{blue}{casual, everyday capture rather than a specific artistic approach}.
        \\
        
        4. Aesthetic Impression Analysis:
        The \textcolor{blue}{tone style is somewhat neutral, lacking vibrancy or warmth} that could enhance the inviting nature of the drink. The \textcolor{blue}{creativity is minimal}, focusing solely on presenting the drink without additional context or storytelling elements. The \textcolor{blue}{environment and background are simple and unobtrusive}, integrating well with the subject. The overall impression is functional but lacks emotional engagement.
        \\
        
        5. Comprehensive Evaluation:
        To improve the image's appeal, \textcolor{blue}{addressing the overexposure and adjusting the color temperature} would significantly enhance the visual experience. \textcolor{blue}{Increasing the contrast and slightly warming the tones} can make the drink more visually appealing. Adjusting the depth of field to create a subtle bokeh effect in the background could further draw attention to the main subject. \textcolor{blue}{Transforming the photography style to something like "Food"} could add a professional touch. \\
        
        \midrule
        \rowcolor{gray!10} \multicolumn{1}{c}{\textbf{Edit Suggestions}} \\ 
        \midrule
        
        1. decrease light\&exposure; \\ 2. change color temperature to slightly warmer \\
        3. increase contrast \\ 4. slightly increase the depth of field (reduce the bokeh effect) \\

        \midrule
        \rowcolor{gray!10} \multicolumn{1}{c}{\textbf{Image Quality Scoring}} \\ 
        \midrule
        
        Predicted Score: 72/100 \\

        \bottomrule
        
    \end{tabularx}
    \label{fig:case2}
\end{figure}

\begin{figure}[h]
    \centering
    % --- 第一部分：横向并排的两张图片 ---
    \begin{subfigure}[b]{0.42\textwidth}
        \centering
        \includegraphics[width=\textwidth]{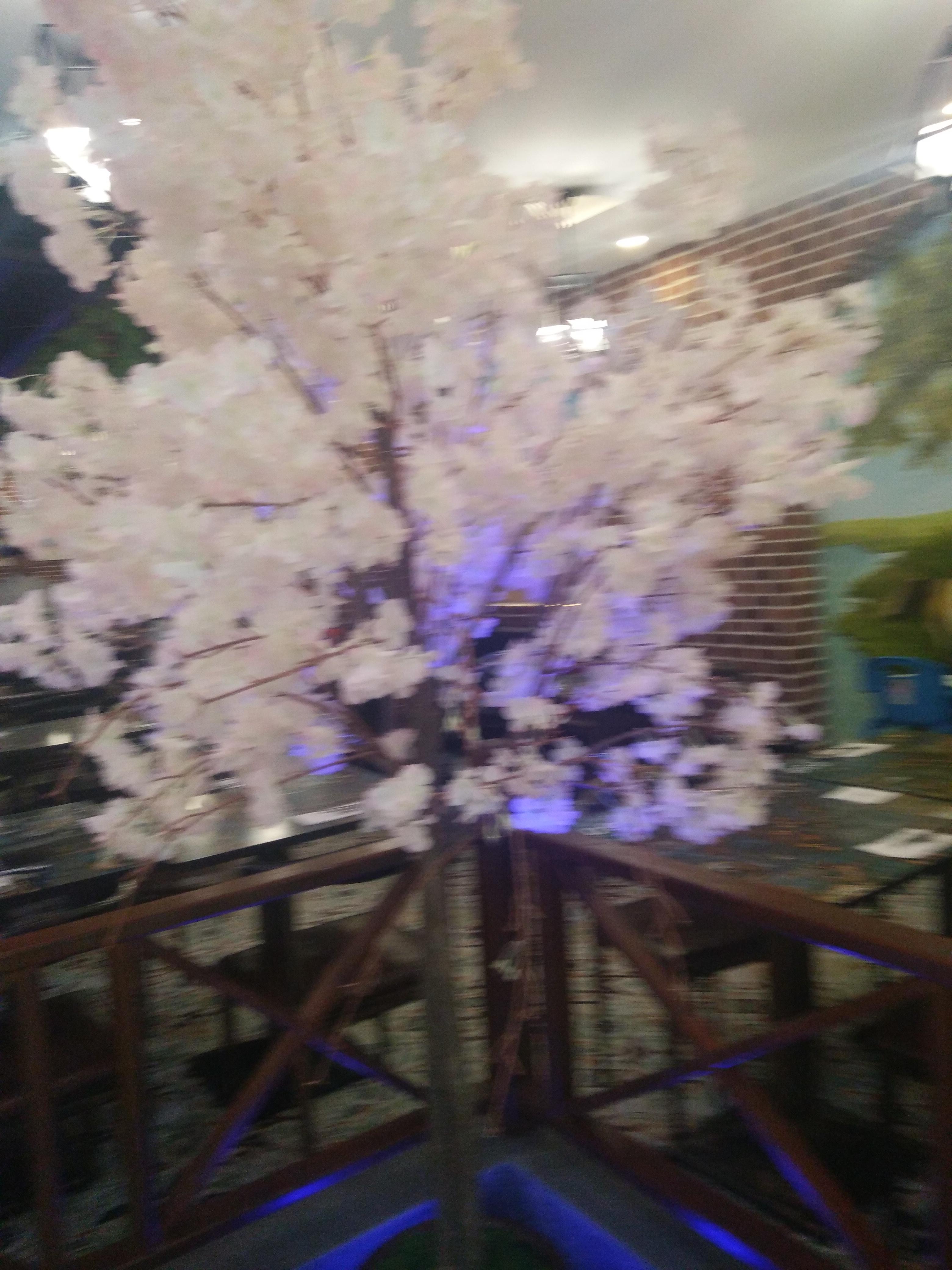} 
        \caption{Input Image}
    \end{subfigure}
    \hfill 
    \begin{subfigure}[b]{0.42\textwidth}
        \centering
        \includegraphics[width=\textwidth]{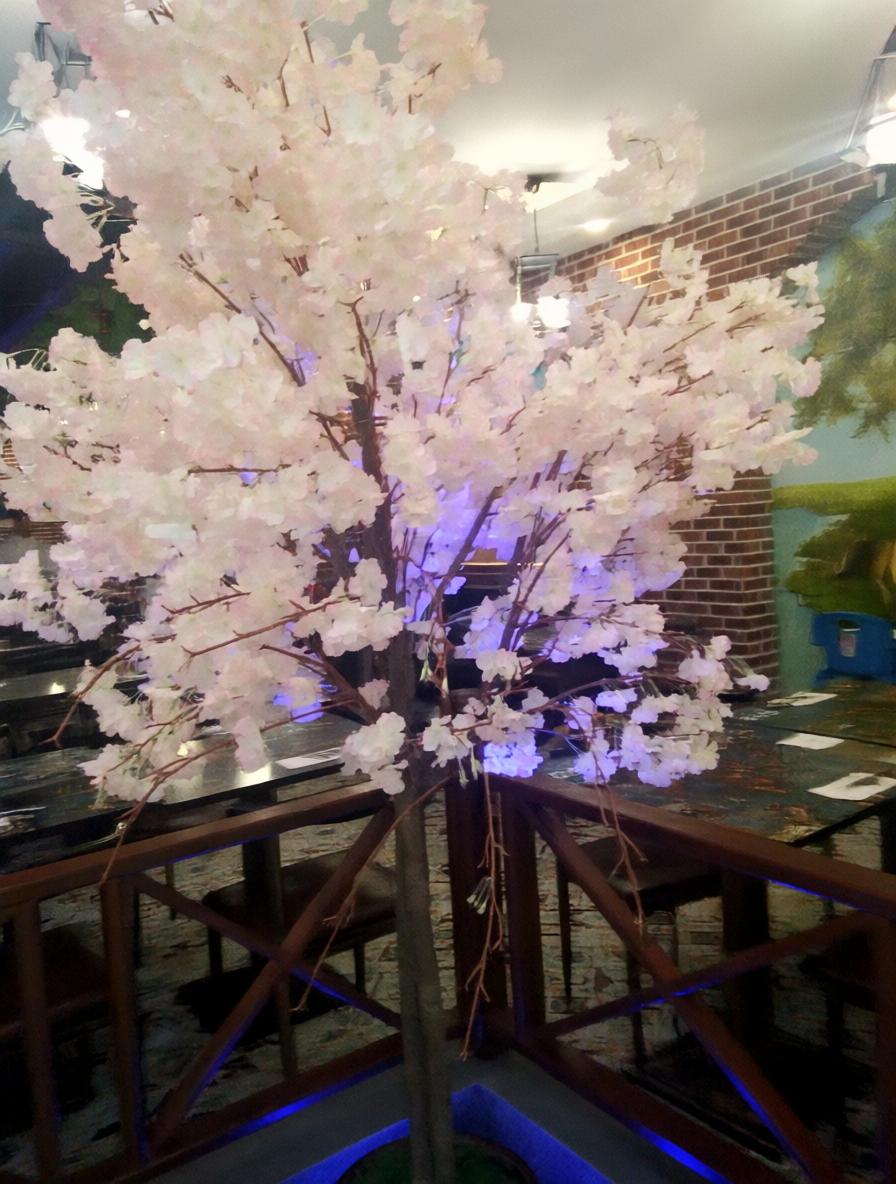} 
        \caption{Enhanced Image}
    \end{subfigure}
    
    % --- 第二部分：图片的总标题 (Case 1. GT Score) ---
    % \vspace{2mm}
    \caption{Ground-truth Score: 37.18}
    
    \vspace{4mm}

    % --- 第三部分：推理结果表格 (无对比，标题居中) ---
    \small
    \begin{tabularx}{\textwidth}{X}
        \toprule
        % \multicolumn{1}{c} 确保标题在 X 列中居中
        \rowcolor{gray!10} \multicolumn{1}{c}{\textbf{Image Quality\&Aesthetic Reasoning}} \\ 
        \midrule
        1. Image Quality/Degradations Analysis: 
        The image suffers from \textcolor{blue}{significant motion blur}, which affects the clarity and sharpness of the entire scene. There is \textcolor{blue}{no noticeable noise or compression artifacts}, but the blur makes it difficult to discern fine details. \textcolor{blue}{No evident distortion or visual interference} is present.
        \\
        2. Color Performance\&Lighting Analysis:
        The colors appear somewhat muted due to the blur, affecting saturation levels. The white balance seems slightly off, leaning towards \textcolor{blue}{cooler tones}, possibly due to artificial lighting. The \textcolor{blue}{contrast is low}, contributing to the lack of depth. The \textcolor{blue}{exposure is adequate}, but the lack of sharpness diminishes the impact of lighting.
        \\
        3. Composition\&Layout Analysis:  
        The main subject appears to be a decorative tree with white blossoms, but the blur obscures its details. The \textcolor{blue}{composition lacks a clear focal point} due to the lack of sharpness. The camera viewpoint seems to be at \textcolor{blue}{eye level}, but the blur detracts from any intended artistic effect. The depth of field is shallow, but this is overshadowed by the motion blur.
        \\
        
        4. Aesthetic Impression Analysis:
        The tone style leans towards a cool, subdued palette, but the \textcolor{blue}{blur prevents the image from conveying a strong emotional expression or storytelling}. The \textcolor{blue}{environment and background are not clearly integrated} due to the lack of detail. The overall impression is one of \textcolor{blue}{unintentional abstraction rather than deliberate artistic choice}.
        \\
        
        5. Comprehensive Evaluation:
           The image suffers from \textcolor{blue}{severe motion blur, which is the primary issue} affecting its quality and aesthetic appeal. The composition lacks a clear focal point, and the intended artistic style is obscured by the blur. To improve the image, \textcolor{blue}{addressing the blur and enhancing the color performance would be essential steps}.  
        \\
        
        \midrule
        \rowcolor{gray!10} \multicolumn{1}{c}{\textbf{Edit Suggestions}} \\ 
        \midrule
        
        1. remove blur \\ 2. slightly increase contrast \\
        3. moderately increase the depth of field (reduce the bokeh effect)
        \\
        
        \midrule
        \rowcolor{gray!10} \multicolumn{1}{c}{\textbf{Image Quality Scoring}} \\ 
        \midrule
        
        Predicted Score: 37/100\\

        \bottomrule
        
    \end{tabularx}
    
    \label{fig:case3}
\end{figure}

\end{document}